\documentclass[10pt,twocolumn,letterpaper]{article}

\usepackage{wacv}
\usepackage{times}
\usepackage{epsfig}
\usepackage{graphicx}
\usepackage{amsmath}
\usepackage{amssymb}
\usepackage{subcaption}
\newcommand\norm[1]{\left\lVert#1\right\rVert}
\usepackage{bbm}
\usepackage{adjustbox}
\usepackage{multirow}
\usepackage[accsupp]{axessibility}  

\def\wacvPaperID{1008} 

\wacvfinalcopy 

\ifwacvfinal
\fi

\ifwacvfinal
\usepackage[breaklinks=true,bookmarks=false]{hyperref}
\else
\usepackage[pagebackref=true,breaklinks=true,colorlinks,bookmarks=false]{hyperref}
\fi

\begin{document}
	
	\title{High Dynamic Range Imaging of Dynamic Scenes with Saturation Compensation but without Explicit Motion Compensation}
	
	\author{Haesoo Chung \qquad\enspace Nam Ik Cho\\
		Department of ECE, INMC, Seoul National University,	Korea\\
		{\tt\small reneeish@ispl.snu.ac.kr, nicho@snu.ac.kr}
	}
	
	\maketitle
	\thispagestyle{empty}
	
	\begin{abstract}
		High dynamic range (HDR) imaging is a highly challenging task since a large amount of information is lost due to the limitations of camera sensors. For HDR imaging, some methods capture multiple low dynamic range (LDR) images with altering exposures to aggregate more information. However, these approaches introduce ghosting artifacts when significant inter-frame motions are present. Moreover, although multi-exposure images are given, we have little information in severely over-exposed areas. Most existing methods focus on motion compensation, i.e., alignment of multiple LDR shots to reduce the ghosting artifacts, but they still produce unsatisfying results. These methods also rather overlook the need to restore the saturated areas. In this paper, we generate well-aligned multi-exposure features by reformulating a motion alignment problem into a simple brightness adjustment problem. In addition, we propose a coarse-to-fine merging strategy with explicit saturation compensation. The saturated areas are reconstructed with similar well-exposed content using adaptive contextual attention. We demonstrate that our method outperforms the state-of-the-art methods regarding qualitative and quantitative evaluations.
	\end{abstract}
	
	\section{Introduction}
	
With the development of high dynamic range (HDR) display, the demand for HDR content is rapidly increasing. HDR content can provide the viewers rich perceptual experiences and also enhance the performance of subsequent computer vision tasks. Since the direct acquisition of HDR images is practically tricky and requires expensive imaging devices, HDR imaging techniques using low dynamic range (LDR) images are drawing considerable attention.
There have been many methods to generate an HDR image from a single LDR input for this reason, where earlier methods just stretched the dynamic range of the LDR input \cite{landis2002production, akyuz2007hdr,  meylan2006reproduction, didyk2008enhancement, banterle2006inverse, banterle2007framework, rempel2007ldr2hdr, huo2014physiological, kovaleski2009high, kovaleski2014high}, and some recent methods learned LDR-to-HDR mapping using convolutional neural networks (CNNs) \cite{eilertsen2017hdr, endoSA2017, marnerides2018expandnet, Yang_2018_CVPR, lee2018deeprecursive, liu2020single}. However, these single-image HDR reconstruction methods usually suffer from information loss in under-exposed or over-exposed areas. Specifically, a large portion of the LDR image content is washed out in a scene with large lighting variations, which is hard to be recovered.  
	
	\begin{figure}[t]
		\centering
		\begin{subfigure}{.45\textwidth}
			\centering
			\includegraphics[width=\textwidth]{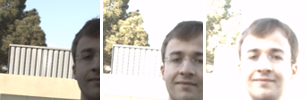}
			\caption{Input LDR images}
			\vspace*{1mm}
			\label{motivation_input}
		\end{subfigure}
		\newline
		\begin{subfigure}{.15\textwidth}
			\includegraphics[width=\textwidth]{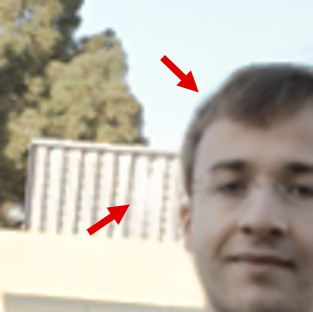}
			\caption{Yan \cite{yan2019attention}}
		\end{subfigure}
		\begin{subfigure}{.15\textwidth}
			\includegraphics[width=\textwidth]{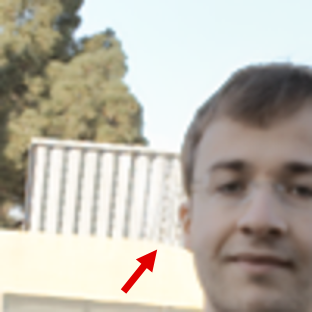}
			\caption{Prabhakar \cite{Prabhakar_2021_CVPR}}
		\end{subfigure}
		\begin{subfigure}{.15\textwidth}
			\includegraphics[width=\textwidth]{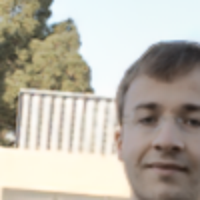}
			\caption{Ours}
		\end{subfigure}
		\caption{Input images with large-scale motions in the over-exposed areas provide insufficient information for HDR reconstruction. Our method successfully hallucinates details in the saturated regions by aggregating similar well-exposed content. The results are visualized after tonemapping.}
		\label{fig:motivation}
	\end{figure} 
		
Hence, there have also been many methods exploiting multi-exposure images to collect more information in dark or very bright areas \cite{debevec1997recovering, mann1994beingundigital, khan2006ghost, grosch2006fast, heo2010ghost, an2011multi, an2012reduction, an2014probabilistic, zhang2012gradient, an2014probabilistic, lee2014ghost, oh2015robust}. But, these methods deliver satisfying performances only when the multi-exposure images are perfectly aligned or have slight movements, which is a hardly realistic situation.

To deal with more dynamic scenes with foreground and background motions, many researchers attempted to align the input images and integrate the aligned LDR images into an HDR result \cite{kang2003high, zimmer2011freehand, kalantari2017deep, sen2012robust, hu2013hdr, prabhakar2016ghosting, kalantari2017deep, wu2018deep, yan2019multi, pu2020robust}. For example, several works \cite{kalantari2017deep, prabhakartowards} used optical flow for aligning input images and passed the aligned images to a fusion network. Wu \etal \cite{wu2018deep} adopted homography transformation for background alignment and the U-Net for HDR reconstruction. Prabhakar \etal \cite{Prabhakar_2021_CVPR} used both homography and optical flow. Meanwhile, Yan \etal \cite{yan2019attention, yan2020deep} utilized attention modules and non-local blocks, respectively, to implicitly align the input features. Most of these methods concentrated on accurate alignment, and then they used relatively simple techniques in merging the aligned LDR images or features. Niu \etal \cite{niu2021hdr} adopted GAN to synthesize missing content but did not perform any explicit hallucination process.
In contrast, we present a coarse-to-fine HDR reconstruction strategy with consideration of the saturated areas. Since details in the over-exposed parts are hardly preserved, we employ the hallucination method. Also, we do not explicitly use optical flow or alignment but transfer the brightness of the multi-exposure images to a reference LDR image so that we can obtain multi-exposure features having the same structure as the reference.
	
More specifically, we propose an end-to-end framework with two sub-networks for HDR imaging of dynamic scenes. First, we present a set of brightness adjustment networks (BANs) that takes the multi-exposure inputs and generates multi-exposure features aligned to those of the reference image. While most of the existing methods transform the pixel position and value of multi-exposure images with respect to a reference, our method transfers the brightness of multi-exposures to the reference to have perfectly aligned multi-exposure images. To this end, each BAN adjusts the exposure of the reference image while retaining its structure using pixel-adaptive deformable convolutions. In addition, we introduce a coarse-to-fine merge-and-hallucination network (MAHN) to integrate the set of multi-exposure features into an HDR image and hallucinate details in the saturated regions. The hallucination is needed because we still have insufficient information in challenging areas, even with the multi-exposure images. For example, when all the images are over-exposed, or occlusions exist in the highlighted areas, naively merging the images leads to poor results, as shown in Fig.~\ref{fig:motivation}. To address this problem, we first coarsely generate an HDR image and then hallucinate content in the saturated regions subsequently. Specifically, we estimate the long-range correlations between the saturated patch and the well-exposed ones and then fill the saturated area with the correlated well-exposed content at the feature level. Extensive quantitative and qualitative evaluations demonstrate that our method generates a high-quality HDR image from LDR images in dynamic scenes. 
	
The main contributions of this paper can be summarized as follows:
\begin{itemize}
\item We propose a brightness adjustment network (BAN) to generate the well-aligned features with different brightness. We reformulate the difficult image-alignment problem into an easier brightness-adjustment problem, which significantly alleviates the ghosting artifacts.
\item We propose a merge-and-hallucination network (MAHN) to integrate the aligned multi-exposure features into an HDR result while hallucinating details in the saturated regions. The MAHN explicitly fills the saturated areas with similar well-exposed content.
\end{itemize}
	
\begin{figure*}[t!]
	\begin{center}
		\includegraphics[width=0.98\linewidth]{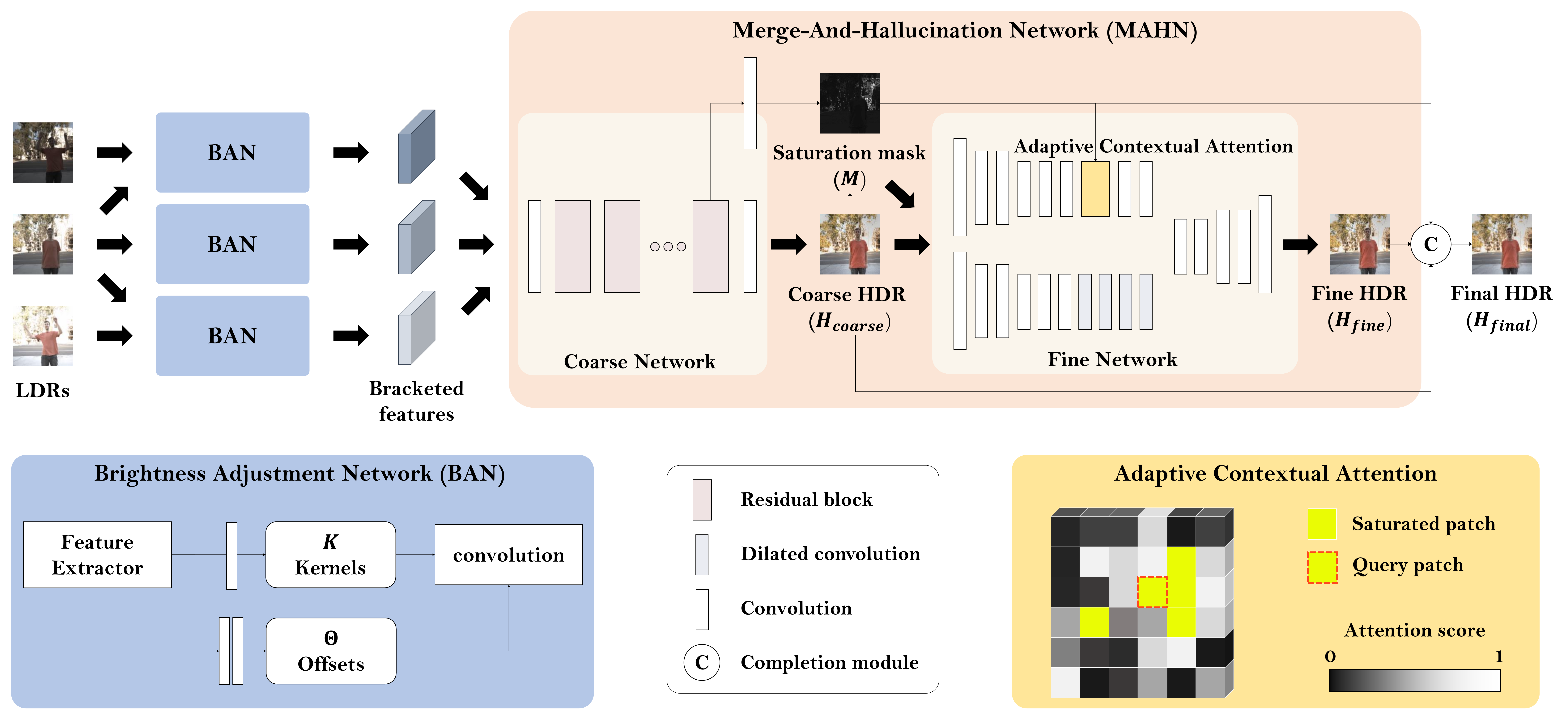}
	\end{center}
	\caption{Overview of the proposed framework. The brightness adjustment network (BAN) adjusts the brightness of the reference image to be matched with the corresponding input. The BAN generates a unique kernel and offset value for each location and applies adaptive convolutions to the reference feature (bottom left). The well-aligned bracketed features obtained from a set of BANs are fed to the merge-and-hallucination network (MAHN). The MAHN first coarsely merges the bracketed features into the coarse HDR image and then hallucinates details in saturated areas using adaptive contextual attention in the fine network. The saturated patches are reconstructed with a weighted sum of the well-exposed content in the adaptive contextual attention layer according to the estimated attention scores. An example of the attention scores for a single query patch is illustrated in the bottom right corner, and the attention for all the saturated patches is computed in the same way.}
	\label{fig:overview}
\end{figure*}
	\section{Related Work}
\noindent{\bf Single-image HDR reconstruction}
Single-image HDR reconstruction, also referred to as reverse tone mapping or inverse tone mapping, has been studied for decades. Early works apply global pixel transformations \cite{landis2002production, akyuz2007hdr}, edit local areas \cite{meylan2006reproduction, didyk2008enhancement}, or utilize an expand map to enhance the highlighted regions \cite{banterle2006inverse, banterle2007framework, rempel2007ldr2hdr}. 
Recently, CNN-based methods \cite{zhang2017learning, marnerides2018expandnet} are introducing end-to-end networks to learn the LDR-to-HDR mapping. Several works \cite{endoSA2017, lee2018deeprecursive} synthesize a multi-exposure stack and combine them into an HDR image, while Eilertsen \etal \cite{eilertsen2017hdr} only restore the saturated regions. More recent approaches \cite{Yang_2018_CVPR, liu2020single} generate an HDR image by reversing the LDR image formation procedure. These methods, however, struggle when severely under-/over-exposed areas exist.
	
\noindent{\bf Multi-image HDR reconstruction}
Multiple images with different exposures can provide richer information for HDR reconstruction. Some conventional methods \cite{debevec1997recovering, mann1994beingundigital} capture a series of LDR images and merge them under the assumption that the scene is static. However, since camera and object motions are inevitable in the real world, the subsequent works propose various methods to handle the displacements. A number of approaches \cite{khan2006ghost, grosch2006fast, heo2010ghost, raman2011reconstruction, zhang2012gradient, an2014probabilistic, lee2014ghost, oh2015robust} assume that input images are globally aligned and focus on detecting and rejecting moving pixels. However, these methods lose considerable information by dropping the pixels with motions. To perform an explicit alignment, several methods \cite{kang2003high, zimmer2011freehand, kalantari2017deep} exploit optical flow, but flow estimation error easily leads to distortions in the resulting image. Meanwhile, Sen \etal \cite{sen2012robust} and Hu \etal \cite{hu2013hdr} rely on patch-based correspondences.
	
Recently, CNN-based multi-exposure HDR imaging methods have been developed. For some examples, Kalantari \etal \cite{kalantari2017deep} compensate for motions using an optical flow algorithm and merge the resulting images using a simple CNN. Wu \etal \cite{wu2018deep} first perform homography transformation to align background motions and pass the aligned images to an image translation network. Yan \etal \cite{yan2019attention} leverage spatial attention to exclude the misaligned components and construct a deep network for merging, while Yan \etal \cite{yan2020deep} use non-local blocks to align the input features. Also, Yan \etal \cite{yan2019multi} and Prabhakar \etal \cite{prabhakartowards} align the images using optical flow and feed them into a fusion network. Pu \etal \cite{pu2020robust} use deformable convolution to align the dynamic input images, and Niu \etal \cite{niu2021hdr} utilize residual merging blocks for alignment and expect the adversarial learning to help to restore missing details.
These approaches mostly give weight to the sophisticated alignment and expect the merging network to combine the aligned features/images well. On the contrary, we present a coarse-to-fine HDR reconstruction strategy with supervision on the saturated regions and a brightness adjustment method to produce well-aligned bracketed features.

\noindent{\bf Flexible convolutions}
Jia \etal \cite{jia2016dynamic} propose a dynamic filter network to generate per-pixel filters conditioned on an input, which has been applied to various tasks dealing with motions \cite{niklaus2017video, mildenhall2018burst, kim2018spatio}. Meanwhile, Dai \etal \cite{dai2017deformable} present deformable convolution to enable flexible operations with learnable offsets. Zhu \etal \cite{zhu2019deformable} extend this work by introducing a modulation factor. Notably, the deformable convolution has been used in various fields related to videos \cite{bertasius2018object, zhao2018trajectory}. Especially, recent video super-resolution methods \cite{tian2020tdan, wang2019edvr, xiang2020zooming, ying2020deformable} leverage deformable convolutions to align the multiple input frames. In this work, we predict spatially varying deformable kernels to deal with a challenging image pair with significant motion and exposure difference.

\section{Proposed Method}
Given a series of LDR images \{$I_{-N}, \dots ,I_0, \dots ,I_N$\} sorted by their exposure biases, our goal is to generate an artifact-free HDR image $H_{final}$. The proposed method consists of two stages, as shown in Fig.~\ref{fig:overview}. First, we place the proposed BANs to control the brightness of the reference image according to other exposure ones. Instead of aligning an exposure image to fit the structure of the reference, we adjust the brightness of the reference with respect to other exposures using the proposed BANs. Each BAN takes the reference and a differently-exposed image and applies pixel-adaptive deformable convolutions to the reference feature to adjust its brightness to that of the different exposure. The resulting multi-exposure features are stacked and fed into the MAHN, which first merges the given features into a coarse HDR image $H_{coarse}$ and then hallucinates details in the saturated regions. The saturated areas are identified by the network and represented as a saturation mask $M$. The fine network then computes contextual attention \cite{yu2018generative} adaptively to find similar content from unsaturated regions and fills the saturated areas with the correlated well-exposed content according to the attention scores. The completion module outputs the final HDR result $H_{final}$ by replacing the saturated parts in the coarse HDR image $H_{coarse}$ with the corresponding ones in the fine HDR image $H_{fine}$.

Following a previous work \cite{kalantari2017deep}, which provides a well-prepared dynamic multi-exposure dataset, we use three LDR images \{$I_{-1}, I_0, I_1$\} and set the middle exposure image $I_0$ as the reference image in terms of structure. 
Here, we use static input images as well as original dynamic input images for training. The static image set \{$I_{-1}^s, I_0^s, I_1^s$\} is generated by adjusting the exposure of the ground truth HDR image $H$ and then applying gamma correction and clipping:
	\begin{equation}
	I_i^s = clip((H t_i)^{1/\gamma}),~i=-1,0,1,
	\end{equation} 
where $t_i$ denotes the exposure time of the corresponding dynamic input image $I_i^d$ and $\gamma$ denotes the gamma correction parameter. $\gamma$ is set as $ 2.2 $ in our experiments. Since the proposed BAN aims to generate the bracketed features instead of learning motions, using static images as inputs does not hinder its training. The static images can serve as easy training samples. 
Then, we map the LDR images to the HDR images \{$H_{-1}, H_0, H_1$\} using gamma correction:
	\begin{equation}
		H_i = \frac{I_i^\gamma}{t_i},~i=-1,0,1,
	\end{equation} 
We concatenate the LDR images with these HDR images along the channel dimension to obtain the $6$-channel inputs \{$X_{-1}, X_0, X_1$\}. Our framework $f$ is represented as follows:
	\begin{equation}
		H_{final} = f(X_{-1}, X_0, X_1).
	\end{equation} 
	\subsection{Brightness Adjustment Network (BAN)}
Given the reference image $I_0$ and the supporting image $I_i$, the BAN aims to adjust the brightness of the reference image $I_0$ in accordance with the exposure of the supporting image $I_i$. To this end, the BAN extracts features to predict spatially-varying deformable convolution kernels and applies the adaptive convolutions to deal with the input pair with motion and brightness difference. Note that we perform self-adjustment in the middle BAN. The generated features are free from the ghosting artifacts since we do not compensate for large motions between the input images but generate the adjusted features of $I_0$ that do not have structure-difference from the reference. 
	
	\noindent\textbf{Feature extraction} 
Our feature extractor has an individual branch for each input and fuses the information from two branches in a progressive manner. The features from the reference image are integrated into the supporting branch so that multi-level information is propagated. The detailed architecture of the feature extractor is illustrated in the supplementary material.
	
	\noindent\textbf{Adaptive convolution} 
The extracted features are then passed to two separate paths to produce convolution kernels $K$ and offsets $\Theta$. The kernels $K$ and the offsets $\Theta$ are unique for each position $\mathbf{p}_0$ on the feature map. Here, we set the kernel size as $3\times3$ and the regular grid as $\mathcal{R}=\{(-1,-1), (-1, 0), \dots , (1, 1)\}$. With the pre-specified offset $\mathbf{p}_n \in \mathcal{R},~n=1,\dots,|\mathcal{R}|$, the adaptive convolution is applied to each location $\mathbf{p}_0$ on the reference feature $F_0$ to generate the adjusted feature $ \bar{F}_{0,i} $ whose brightness matches with the one of the supporting feature $F_i$:
\begin{equation}
\bar{F}_{0,i}(\mathbf{p}_0)=\sum_{\mathbf{p}_n \in \mathcal{R}}K(\mathbf{p}_0+\mathbf{p}_n)F_0(\mathbf{p}_0+\mathbf{p}_n+\Delta \mathbf{p}_n),
\end{equation}
where $\Delta\mathbf{p}_n \in \Theta$ is the learnable offset. Since $\mathbf{p}_0+\mathbf{p}_n+\Delta \mathbf{p}_n$ can be fractional, bilinear interpolation is used to compute the value $F_0(\mathbf{p}_0+\mathbf{p}_n+\Delta \mathbf{p}_n)$. 
	
Unlike previous works \cite{tian2020tdan, wang2019edvr, xiang2020zooming, ying2020deformable} which adopt the deformable convolution, the proposed method applies the convolutions to the reference feature, and the supporting image is only involved in feature extraction. We prevent undesirable ghosting artifacts by converting the alignment problem into an easier brightness adjustment problem. We show the effect of this reformulation in Section \ref{ablation}. Furthermore, learning per-pixel kernels as well as offsets enables more flexible operations.
\subsection{Merge-And-Hallucination Network (MAHN)}
For generating a high-quality HDR result, the MAHN first merges the well-aligned bracketed features into a coarse HDR image $H_{coarse}$ and then hallucinates details within the saturated areas by aggregating similar content in the unsaturated (\textit{i.e.}, well-exposed) areas. The coarse network integrates the bracketed features using residual blocks. The following fine network is branched into two paths: a hallucination branch and a refinement branch. While the refinement branch refines the coarse result with dilated convolutions locally, the hallucination branch explicitly fills the saturated regions with similar valid content by estimating long-range adaptive contextual attention.
	
\begin{figure}[t]
	\centering
	\centering
	\begin{subfigure}{.135\textwidth}
		\includegraphics[width=\textwidth]{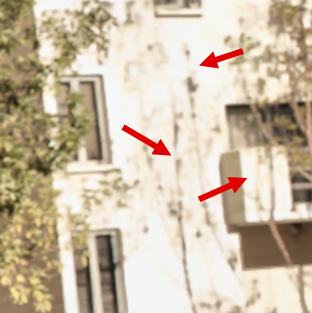}
		\caption*{$H_{coarse}$}
	\end{subfigure}
	\begin{subfigure}{.135\textwidth}
		\includegraphics[width=\textwidth]{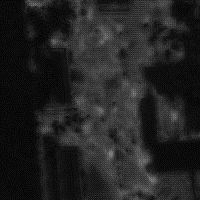}
		\caption*{$M$}
	\end{subfigure}
	\begin{subfigure}{.135\textwidth}
		\includegraphics[width=\textwidth]{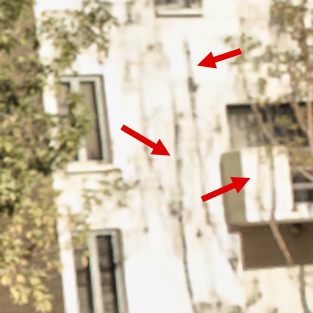}
		\caption*{$H_{fine}$}
	\end{subfigure}
	\caption{The saturation mask $M$ represents the saturation level of the coarse HDR image $H_{coarse}$. As a result of hallucination using the adaptive contextual attention, the fine HDR image $H_{fine}$ clearly retains richer details in the saturated parts.}
	\label{fig:method_ca}
\end{figure}

Our core idea is to find the correlated valid content to the saturated parts and assemble them to compensate for the lost content in the saturated areas. First, we construct a saturation mask $M$ to indicate the saturated regions. 
We place a convolutional layer with a  sigmoid activation function after the last residual block in the coarse network to obtain the sample-specific saturation mask. The saturation mask $M$ represents how saturated each pixel is with values in range $ [0,1] $. An example is shown in Fig.~\ref{fig:method_ca}. This adaptive mask generation strategy enables sample-specific hallucination, contrary to pre-defined rules such as thresholding. 
The hallucination branch fills the over-exposed regions (close to value $1$ on the $M$) with the correlated well-exposed content (close to value $0$ on the $M$) by measuring the patch-wise cosine similarity:
\begin{equation}
s_{ x,y,x',y' } = \frac{o_{x,y}}{\norm{o_{x,y}}} \cdot \frac{w_{x',y'}}{\norm{w_{x',y'}}} ,
\end{equation}
where $o_{x,y}$ denotes the over-exposed patch at $(x,y)$ and $w_{x,y}$ denotes the well-exposed patch at $(x',y')$. To obtain the attention scores, we apply softmax along the $x'y'$-dimension and then multiply $ 1-M $ which represents well-exposedness so that valid patches can contribute more to the reconstruction. We replace the over-exposed patches with a combination of the well-exposed patches according to the estimated attention scores. The effectiveness of our adaptive contextual attention is validated in Section \ref{ablation}. Note that the whole process is implemented using convolution operations.
	
The saturation-compensated features are concatenated with the refined features from the refinement branch and go through additional layers to reconstruct the fine HDR image $H_{fine}$. The fine HDR image $H_{fine}$ exhibits clear improvements in the saturated areas as shown in Fig.~\ref{fig:method_ca}.
Finally, the completion module generates the final HDR image $H_{final}$ by replacing the saturated pixels of $H_{coarse}$ with the ones of $H_{fine}$:
\begin{equation}
H_{final} = (1-M) \odot H_{coarse} + M \odot H_{fine},
\end{equation}
where $\odot$ denotes the Hadamard product.

	\subsection{Training Loss}
We propose a hybrid loss to enhance both fidelity and perceptual quality. Since HDR images are mostly tonemapped for displaying, we compute the loss functions between the tonemapped predicted HDR image $\mathcal{T}(\hat{H})$ and the tonemapped ground truth HDR image $\mathcal{T}(H)$. The HDR image $H$ is tonemapped with the differentiable $\mu$-law: 
\begin{equation}
	\mathcal{T}(H) = \frac{\log(1+\mu H)}{\log(1+H)},
\end{equation}
where $\mu$ is a parameter that defines the level of compression. $\mu$ is set as $ 5000 $.
	
\noindent\textbf{Reconstruction loss}
We use a simple $\ell_1$ reconstruction loss to minimize the distance between $\mathcal{T}(\hat{H})$ and $\mathcal{T}(H)$. The reconstruction loss is defined as:
\begin{equation}
\mathcal{L}_{recon}=\norm{\mathcal{T}(\hat{H}) - \mathcal{T}(H)}_1.
\end{equation}

\noindent\textbf{Color loss}
To address the color shift problem, we also define color loss, which is based on the cosine similarity between the RGB vectors of $\mathcal{T}(\hat{H})$ and $\mathcal{T}(H)$. Formally, it is described as:
	\begin{equation}
		\mathcal{L}_{color}=1 - \frac{1}{N}\sum_{n=1}^{N}\frac{\hat{\textbf{v}}_n \cdot \textbf{v}_n}{\norm{\hat{\textbf{v}}_n} \norm{\textbf{v}_n}},
	\end{equation}
	where $N$ is the total number of pixels of the HDR image, and $\hat{\textbf{v}}_n$ and $\textbf{v}_n$ denote the RGB vectors at the $n$-th pixel of $\mathcal{T}(\hat{H})$ and $\mathcal{T}(H)$, respectively.

\noindent\textbf{Perceptual loss}
To generate a more realistic texture, we adopt the VGG loss $\mathcal{L}_{vgg}$ as our perceptual loss. We use three feature maps after the first, second, and third block of VGG-16 network for the VGG loss. The total variation (TV) loss $\mathcal{L}_{tv}$ is also included for smoothness.

\noindent\textbf{Total loss}
From the above losses, the overall loss is defined as $ \mathcal{L}= \lambda_{recon}\mathcal{L}_{recon} + \lambda_{color}\mathcal{L}_{color} + \lambda_{vgg}\mathcal{L}_{vgg} + \lambda_{tv}\mathcal{L}_{tv}, $
which is applied to the coarse output $H_{coarse}$, the fine output $H_{fine}$, and the final output $H_{final}$ with different weights. We empirically set the corresponding weights as specified in Table~\ref{tab:loss}. 

\begin{table}[h!]
	\caption{The weights for each loss constituting our total loss.}
	\centering
	\label{tab:loss}
	\begin{tabular}{l|cccc}
		\hline
		& $\lambda_{recon}$ & $\lambda_{color}$ & $\lambda_{vgg}$ & $\lambda_{tv}$   \\ 
		\hline
		$H_{coarse}$ & 1               & 1               & 0.001           & 0.1  \\
		$H_{fine}$   & 1               & 1               & 0.001           & 0.1  \\
		$H_{final}$  & 1               & 0               & 0             & 0      \\
		\hline
	\end{tabular}
\end{table}

	
	\subsection{Implementation Details}
We sample patches of size $ 128\times128 $ from the training images and apply augmentations: random rotation and flipping. We use Adam optimizer with a learning rate of $10^{-4}$ and set the batch size as $16$. Each batch is composed of the dynamic images and the static images with the ratio of $3:1$. We train our model on a single NVIDIA RTX 2080 Ti GPU.

\begin{table*}[t!]
	\caption{Quantitative comparisons of our method with state-of-the-art methods. \textsuperscript{$\dagger$} indicates that the values are taken from their original papers. O.F. and Homo. refer to the optical flow-based alignment and the homography transformation, respectively.}
	\label{tab:comparisons}
	\centering 
	\begin{tabular}{lcccccccc}
		\hline\\[-1em]
		\multirow{2}{*}{Methods}                  & \multicolumn{2}{c}{Pre-alignment} & \multicolumn{1}{c}{Boundary} & \multirow{2}{*}{PSNR\textsubscript{$T$}} & \multirow{2}{*}{SSIM\textsubscript{$T$}} & \multirow{2}{*}{PSNR\textsubscript{$L$}} & \multirow{2}{*}{SSIM\textsubscript{$L$}} & \multirow{2}{*}{HDR-VDP-2} \\ \cline{2-3}\\[-1em]
		& ~~O.F.~~              & Homo.              & \multicolumn{1}{c}{Cropping} &                    &                    &                    &                    &                    \\ \hline\\[-1em]
		Sen \cite{sen2012robust} & & & & 41.11	        & 0.9815          & 38.82     	   & 0.9749        	 & 57.43         \\
		Hu \cite{hu2013hdr}    & & & & 34.87     	    & 0.9698     	  & 31.72    	   & 0.9511          & 55.20         \\
		AHDRNet \cite{yan2019attention}     &  & & & 42.22          & 0.9904      	  & 41.26 & 0.9862          & 61.54         \\ 	
		NHDRRNet\textsuperscript{$\dagger$} \cite{yan2020deep}    &   & & & 42.41          & 0.9887      	  & - & -          & -        \\ 			
		Prabhakar\textsuperscript{$\dagger$} \cite{prabhakartowards}    &$\checkmark$  & & & 42.82          & -      	  & 41.33 & -          & -        \\ 
		HDR-GAN\textsuperscript{$\dagger$} \cite{niu2021hdr}     &  & & & 43.92          & 0.9905      	  & 41.57 & 0.9865          & -         \\ 							
		Ours      		 & & & & \textbf{44.48}  		&\textbf{0.9917} & \textbf{42.45}          & \textbf{0.9880} & \textbf{61.76} \\ \hline\\[-1em]
		Kalantari \cite{kalantari2017deep}&$\checkmark$  & &$\checkmark$ & 42.83 & 0.9877      	  & 41.49          & 0.9858          & 59.82         \\
		Ours      		 & & &$\checkmark$ & \textbf{43.42}  		&\textbf{0.9892} & \textbf{41.68}          & \textbf{0.9866} & \textbf{61.81} \\ \hline\\[-1em]			
		Wu \cite{wu2018deep}     &  &$\checkmark$ &$\checkmark$ & 42.49          & 0.9889      	  & 42.06 & 0.9870          & 61.30         \\ 	
		Prabhakar \cite{Prabhakar_2021_CVPR}      &$\checkmark$  &$\checkmark$ &$\checkmark$ & 41.95          & 0.9873      	  & 41.82 & \textbf{0.9879}          & 61.23         \\ 	
		Ours      		 & &$\checkmark$ &$\checkmark$ & \textbf{43.11}  		&\textbf{0.9901} & \textbf{42.37}          & \textbf{0.9879} & \textbf{61.70} \\
		\hline
	\end{tabular}%
\end{table*}


	\section{Experiments}
\subsection{Experimental Settings}
\noindent\textbf{Datasets} We use the dataset constructed by Kalantari \etal \cite{kalantari2017deep} for both training and testing. This dataset consists of 74 scenes for training and 15 scenes for testing, each of which contains three dynamic LDR images with different exposures. We also conduct qualitative evaluations on Sen \etal's \cite{sen2012robust} dataset. Both datasets contain LDR images which have large-scale motions and severe saturation.

\noindent\textbf{Evaluation metrics} We use five evaluation metrics for the quantitative evaluation. We compute the PSNR and SSIM values between the predicted and ground truth HDR images after tonemapping (PSNR\textsubscript{$T$} and SSIM\textsubscript{$T$}) and in the linear domain (PSNR\textsubscript{$L$} and SSIM\textsubscript{$L$}). We also calculate the HDR-VDP-2 score \cite{mantiuk2011hdr} to measure the visual quality of HDR images.
	
	\subsection{Comparisons}
We compare our results with previous state-of-the-art methods, including two patch-based methods \cite{sen2012robust,hu2013hdr} and seven CNN-based approaches \cite{kalantari2017deep, wu2018deep, yan2019attention, yan2020deep, prabhakartowards, niu2021hdr, Prabhakar_2021_CVPR}. Note that Kalantari \etal \cite{kalantari2017deep} and Prabhakar \etal \cite{prabhakartowards} first align the input images using optical flow and Wu \etal \cite{wu2018deep} apply homography transformation. Also, Prabhakar \etal \cite{Prabhakar_2021_CVPR} use both of them for pre-alignment. 
We used the official codes if they are provided. Otherwise, we re-implemented their methods according to their papers except three methods \cite{yan2020deep,prabhakartowards,niu2021hdr}. We used the quantitative results reported in their papers, since we could not reproduce their results due to absence of the necessary data \cite{prabhakartowards} or insufficient implementation details \cite{yan2020deep,niu2021hdr}. HDR-VDP-2 score is not taken because it changes depending on the evaluation setting, which is not specified in their papers.

\noindent\textbf{Quantitative evaluations} We compute five metrics mentioned above in Table~\ref{tab:comparisons} for the quantitative evaluations. 
Note that Kalantari \etal's method \cite{kalantari2017deep} needs to crop $6$ pixels near image boundary, thus we compare with this method separately after boundary cropping to evaluate on the same input. Also, the methods of Wu \etal \cite{wu2018deep} and Prabhakar \etal \cite{Prabhakar_2021_CVPR} lose boundary content irregularly due to homography transformation, thus we apply homography transformation and pass the cropped images to our framework for fair comparisons with these methods. But, we do not use homography transformation during training. The evaluations are conducted on full images without losing image boundary for our method and the other six methods \cite{sen2012robust,hu2013hdr,yan2019attention,yan2020deep,prabhakartowards,niu2021hdr}. 
It can be seen that the proposed method achieves the best performance in terms of all metrics, which validates that the results produced by our method are visually pleasing both in the linear domain and after tonemapping. Note that our method is an end-to-end framework which does not require any pre-alignment process such as homography transformation and optical flow algorithm.

\begin{figure*}[t]
	\begin{subfigure}{.085\textwidth}
		\includegraphics[height=3.2cm]{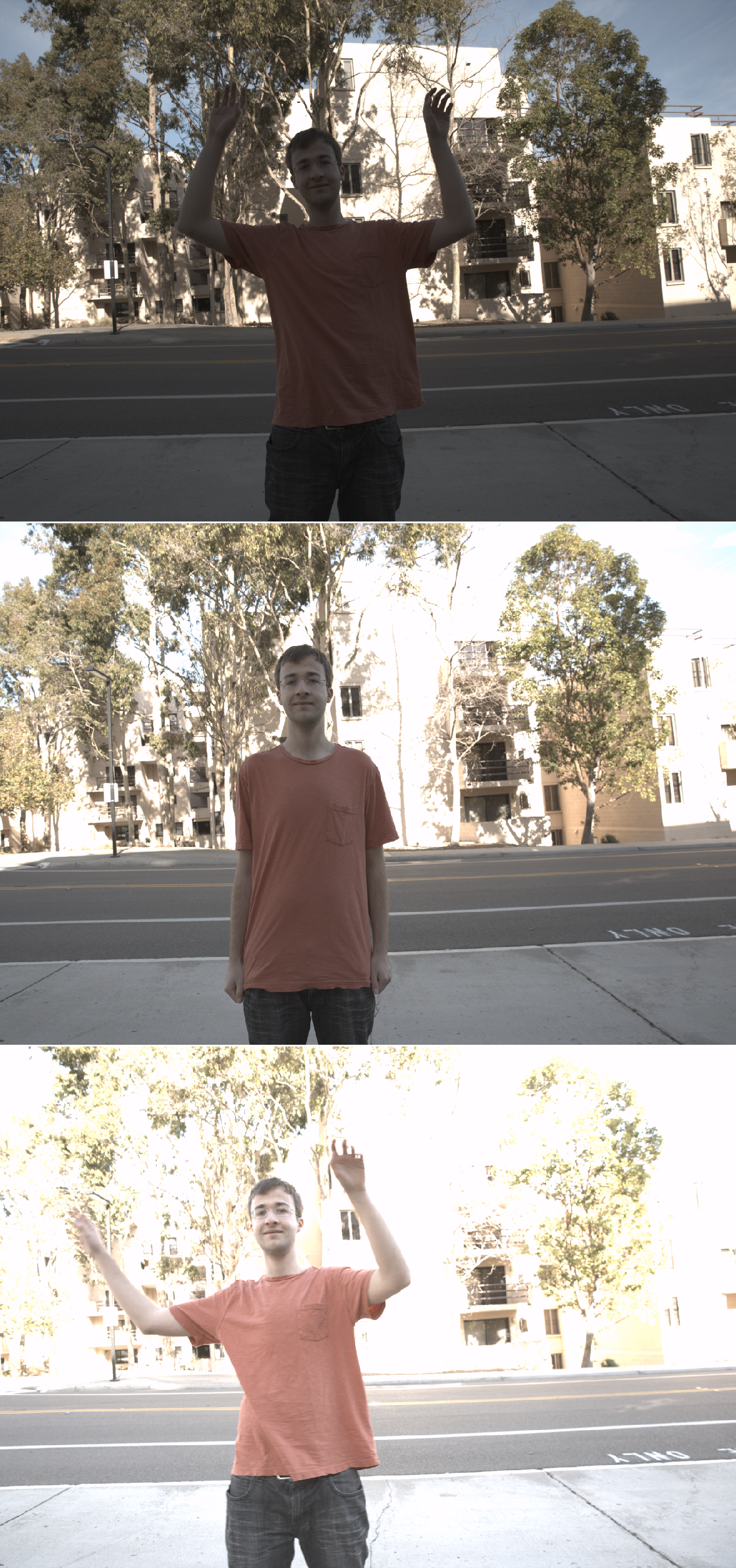}\vspace*{-1mm} \caption*{Input LDRs}
	\end{subfigure}
	\begin{subfigure}{.258\textwidth}
		\includegraphics[height=3.2cm]{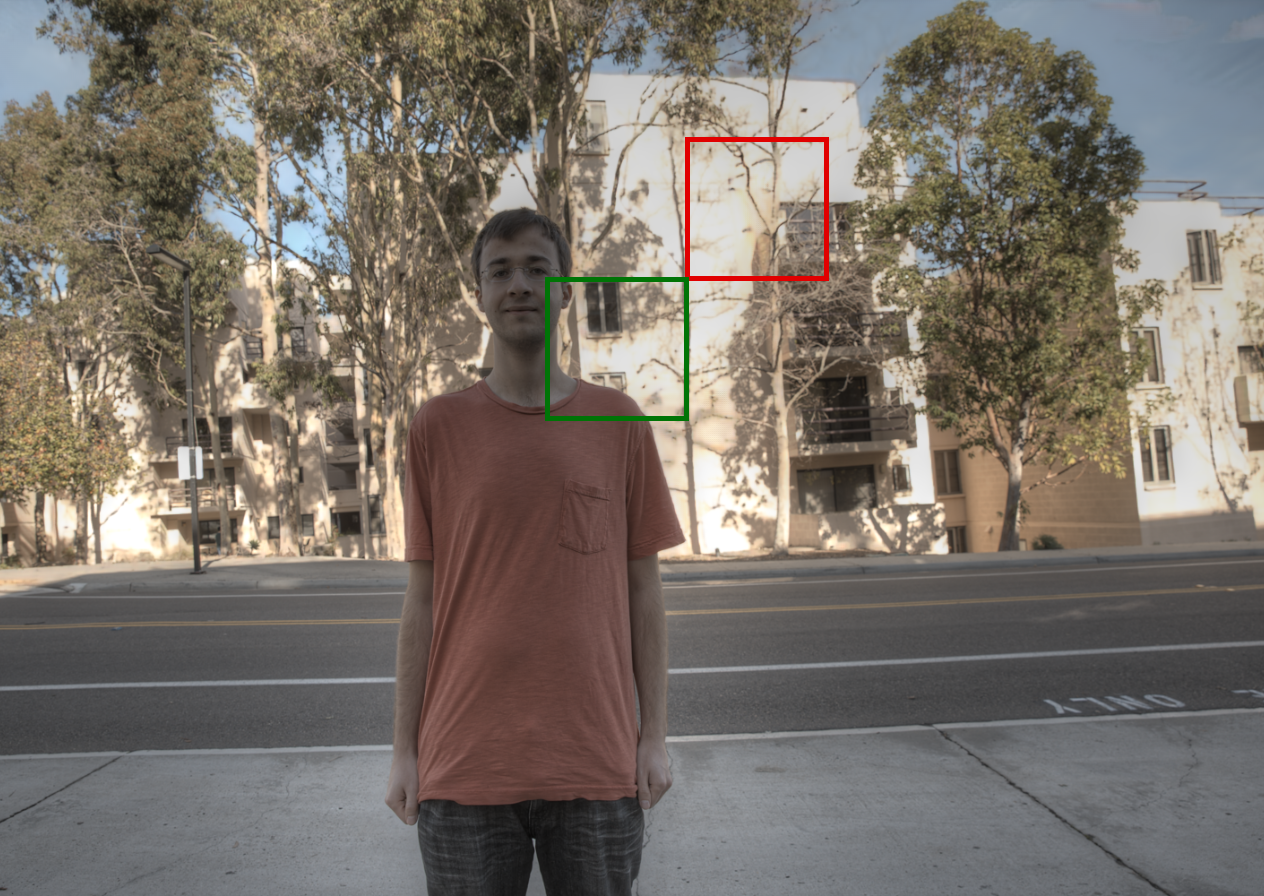}\vspace*{-1mm} \caption*{Our result}
	\end{subfigure}
	\begin{subfigure}{.06\textwidth}
		\includegraphics[height=3.2cm]{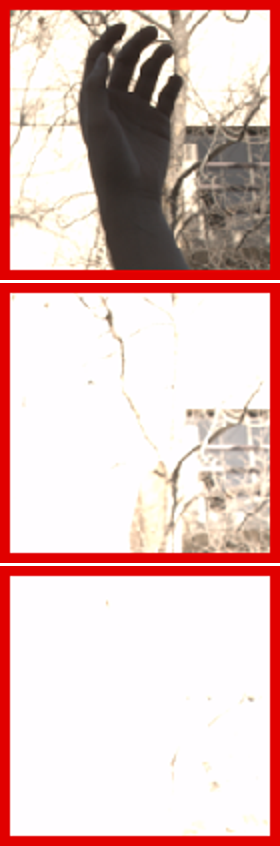}\vspace*{-1mm} \caption*{~\quad LDR}
	\end{subfigure}
	\begin{subfigure}{.06\textwidth}
		\includegraphics[height=3.2cm]{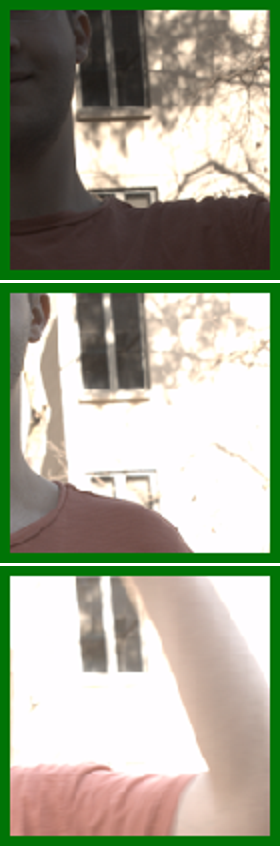}\vspace*{-1mm} \caption*{patches~\quad}
	\end{subfigure}	
	\begin{subfigure}{.085\textwidth}
		\includegraphics[height=3.2cm]{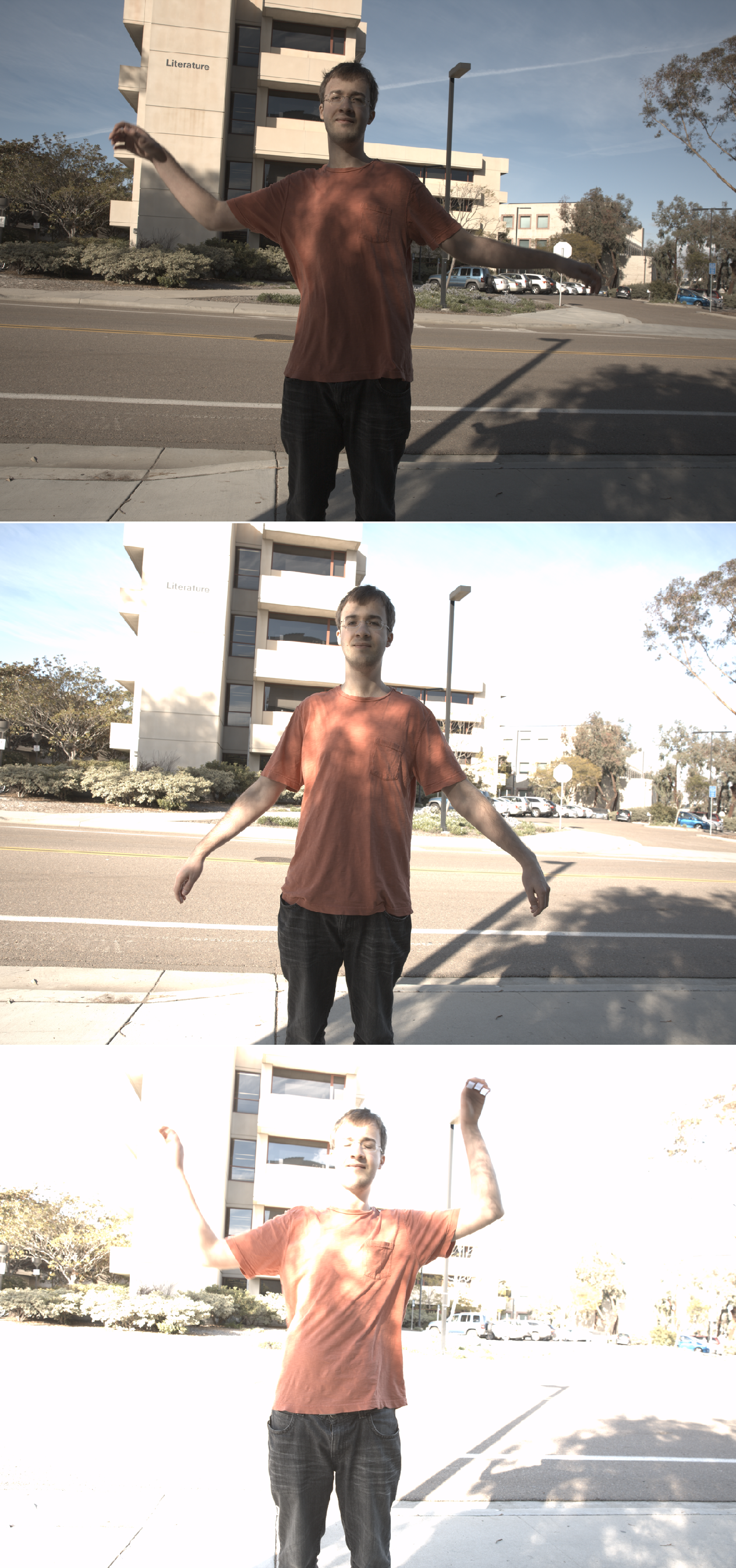}\vspace*{-1mm} \caption*{Input LDRs}
	\end{subfigure}
	\begin{subfigure}{.258\textwidth}
		\includegraphics[height=3.2cm]{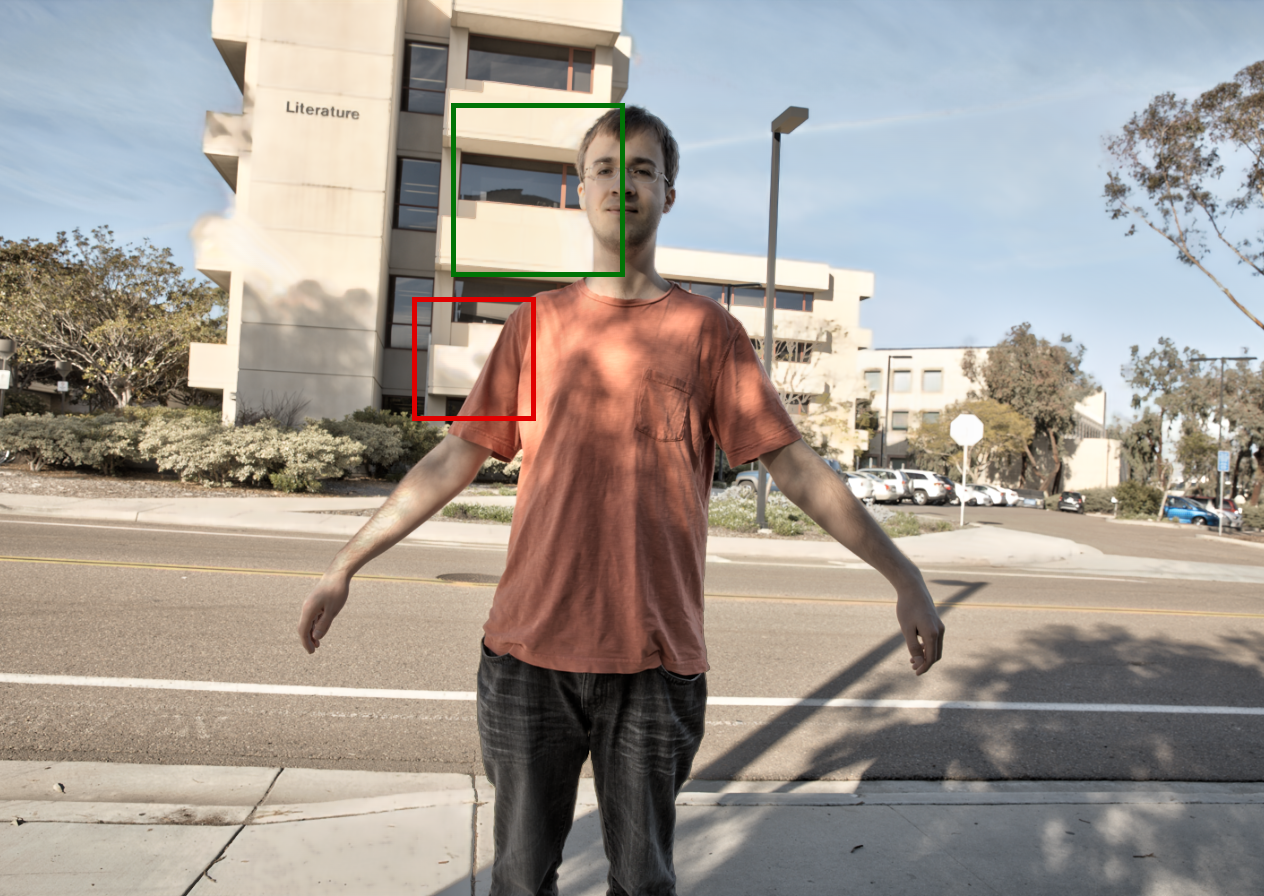}\vspace*{-1mm} \caption*{Our result}
	\end{subfigure}
	\begin{subfigure}{.06\textwidth}
		\includegraphics[height=3.2cm]{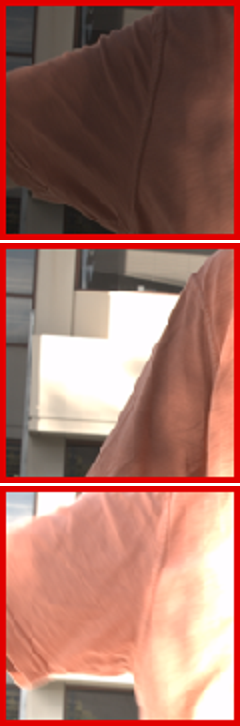}\vspace*{-1mm} \caption*{~\quad LDR}
	\end{subfigure}
	\begin{subfigure}{.055\textwidth}
		\includegraphics[height=3.2cm]{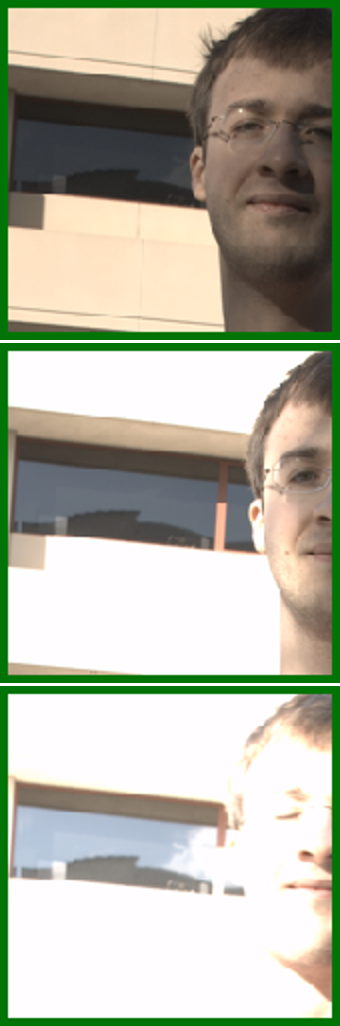}\vspace*{-1mm} \caption*{patches~\quad}
	\end{subfigure}

	\vspace*{1mm}
	\begin{subfigure}{.0554\textwidth}
		\includegraphics[height=1cm]{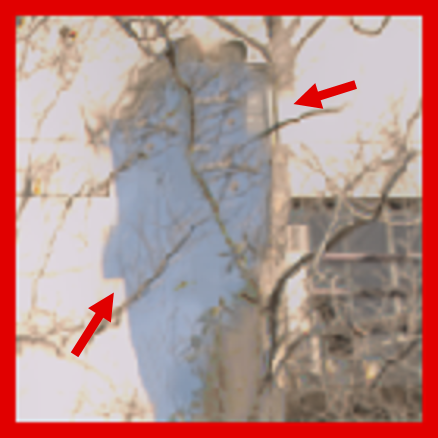}
	\end{subfigure}	
	\begin{subfigure}{.0554\textwidth}
		\includegraphics[height=1cm]{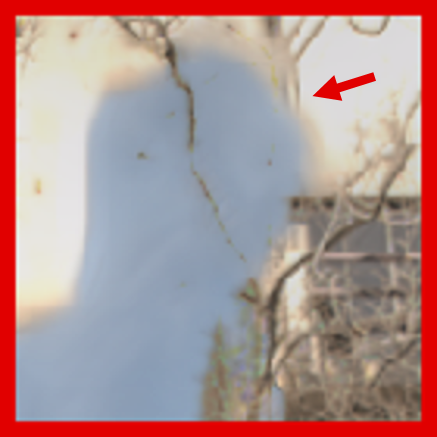}
	\end{subfigure}	
	\begin{subfigure}{.0554\textwidth}
		\includegraphics[height=1cm]{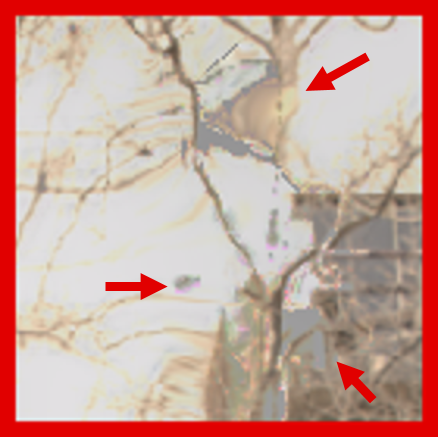}
	\end{subfigure}	
	\begin{subfigure}{.0554\textwidth}
		\includegraphics[height=1cm]{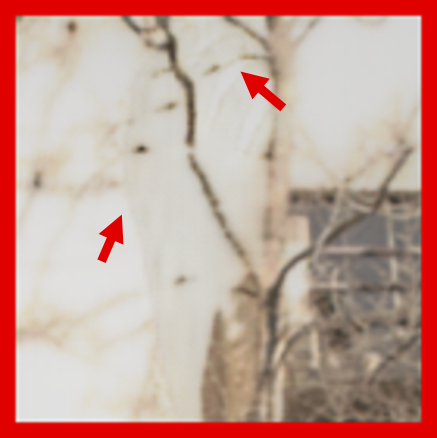}
	\end{subfigure}	
	\begin{subfigure}{.0554\textwidth}
		\includegraphics[height=1cm]{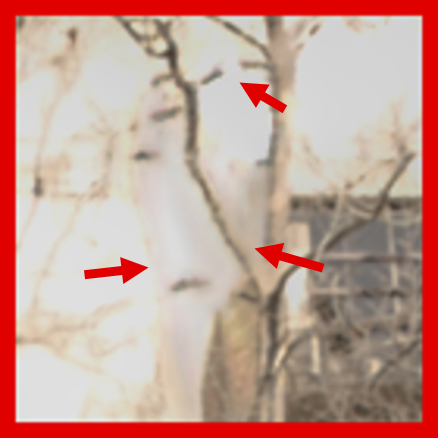}
	\end{subfigure}	
	\begin{subfigure}{.0554\textwidth}
		\includegraphics[height=1cm]{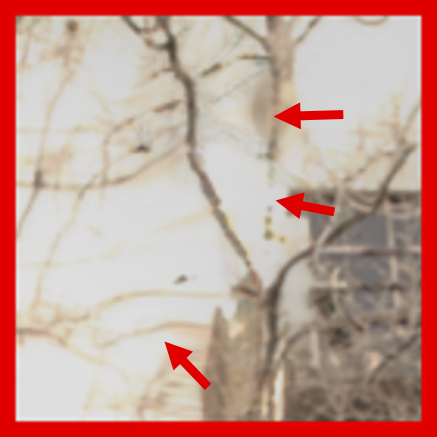}
	\end{subfigure}	
	\begin{subfigure}{.0554\textwidth}
		\includegraphics[height=1cm]{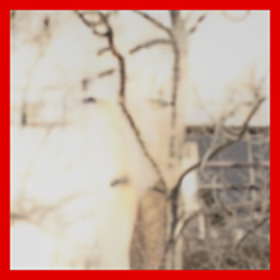}
	\end{subfigure}	
	\begin{subfigure}{.0554\textwidth}
		\includegraphics[height=1cm]{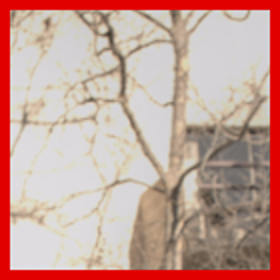}
	\end{subfigure}	
	\begin{subfigure}{.0554\textwidth}
		\includegraphics[height=1cm]{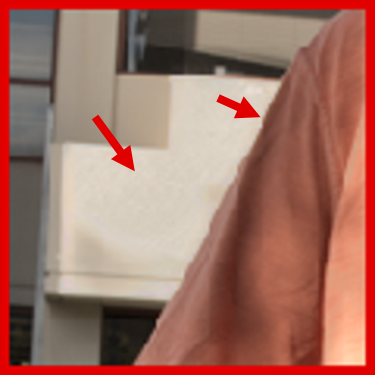}
	\end{subfigure}	
	\begin{subfigure}{.0554\textwidth}
		\includegraphics[height=1cm]{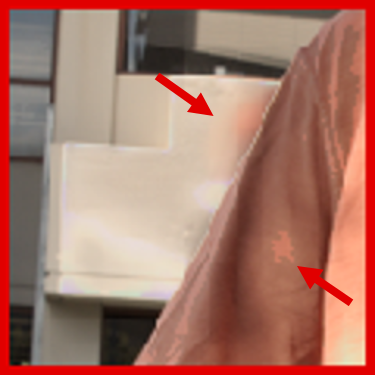}
	\end{subfigure}	
	\begin{subfigure}{.0554\textwidth}
		\includegraphics[height=1cm]{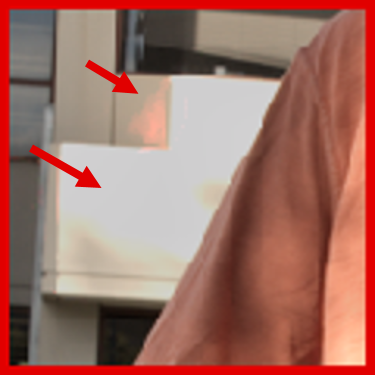}
	\end{subfigure}	
	\begin{subfigure}{.0554\textwidth}
		\includegraphics[height=1cm]{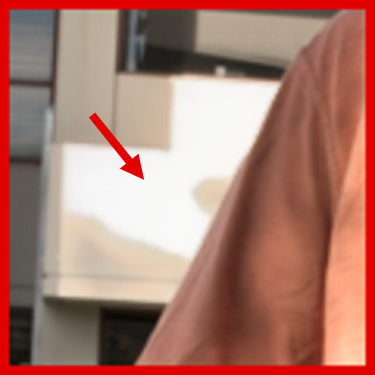}
	\end{subfigure}	
	\begin{subfigure}{.0554\textwidth}
		\includegraphics[height=1cm]{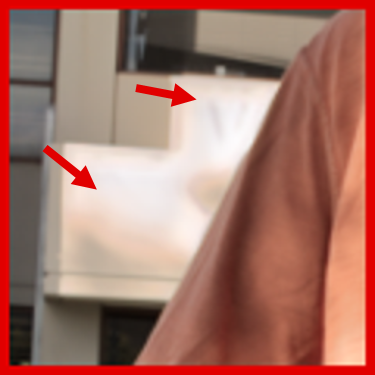}
	\end{subfigure}	
	\begin{subfigure}{.0554\textwidth}
		\includegraphics[height=1cm]{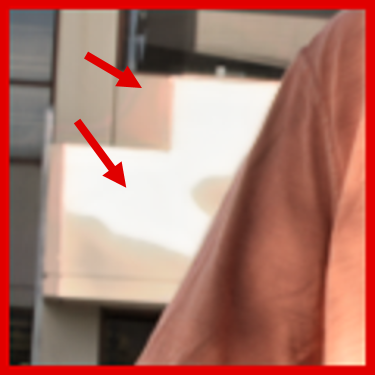}
	\end{subfigure}	
	\begin{subfigure}{.0554\textwidth}
		\includegraphics[height=1cm]{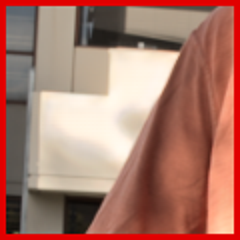}
	\end{subfigure}	
	\begin{subfigure}{.0554\textwidth}
		\includegraphics[height=1cm]{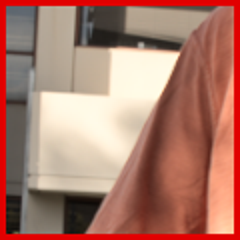}
	\end{subfigure}	
	
	\begin{subfigure}{.0554\textwidth}
		\includegraphics[height=1cm]{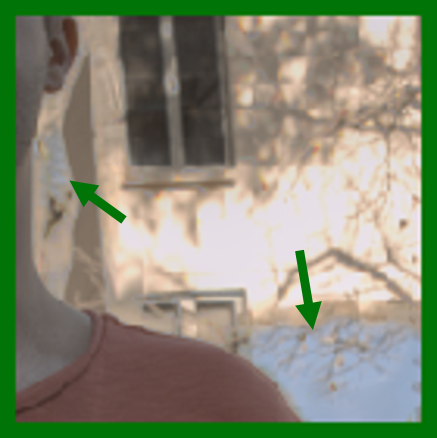} \vspace*{-6mm}
		\caption*{\tiny Sen} \vspace*{-1.5mm}
		\caption*{\footnotesize \cite{sen2012robust}}
	\end{subfigure}	
	\begin{subfigure}{.0554\textwidth}
		\includegraphics[height=1cm]{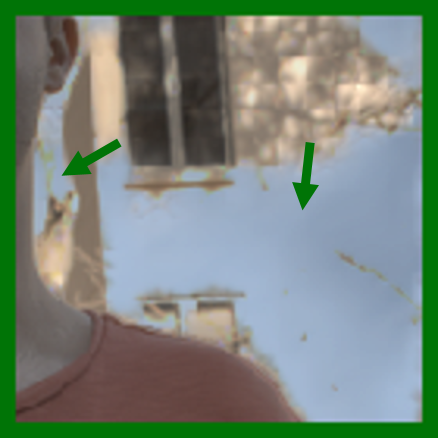} \vspace*{-6mm}
		\caption*{\tiny Hu} \vspace*{-1.5mm}
		\caption*{\footnotesize \cite{hu2013hdr}}
	\end{subfigure}	
	\begin{subfigure}{.0554\textwidth}
		\includegraphics[height=1cm]{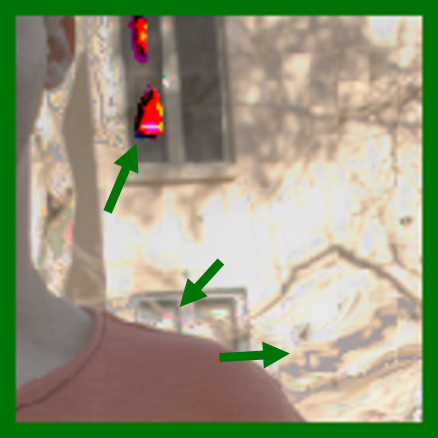} \vspace*{-6mm}
		\caption*{\tiny Kalantari} \vspace*{-1.5mm}
		\caption*{\footnotesize \cite{kalantari2017deep}}
	\end{subfigure}	
	\begin{subfigure}{.0554\textwidth}
		\includegraphics[height=1cm]{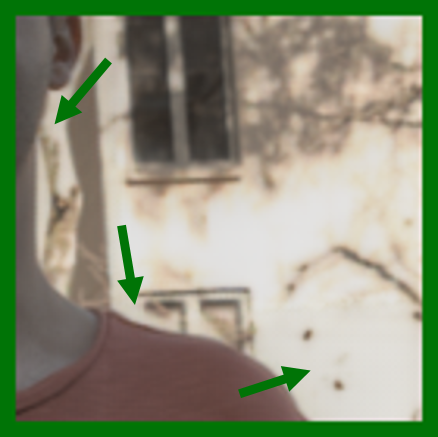} \vspace*{-6mm}
		\caption*{\tiny Wu} \vspace*{-1.5mm}
		\caption*{\footnotesize \cite{wu2018deep}}
	\end{subfigure}	
	\begin{subfigure}{.0554\textwidth}
		\includegraphics[height=1cm]{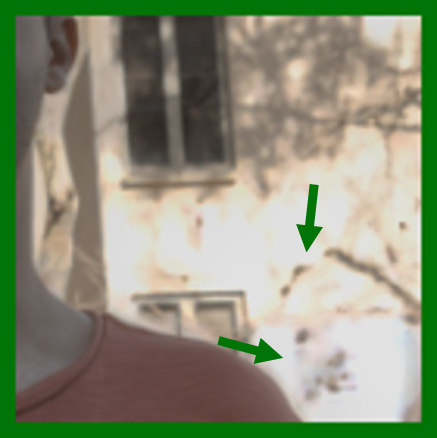} \vspace*{-6mm}
		\caption*{\tiny Yan} \vspace*{-1.5mm}
		\caption*{\footnotesize \cite{yan2019attention}}
	\end{subfigure}	
	\begin{subfigure}{.0554\textwidth}
		\includegraphics[height=1cm]{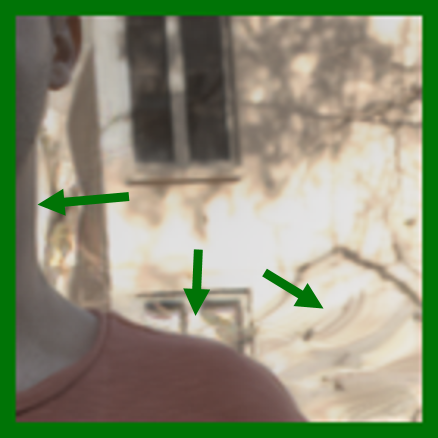} \vspace*{-6mm}
		\caption*{\tiny Prabhakar} \vspace*{-1.5mm}
		\caption*{\footnotesize \cite{Prabhakar_2021_CVPR}}
	\end{subfigure}	
	\begin{subfigure}{.0554\textwidth}
		\includegraphics[height=1cm]{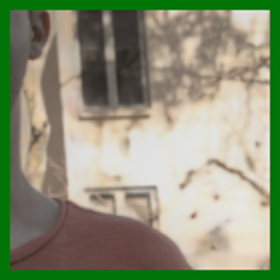} \vspace*{-6mm}
		\caption*{\tiny Ours} \vspace*{-1.5mm}
		\caption*{\footnotesize }
	\end{subfigure}	
	\begin{subfigure}{.0554\textwidth}
		\includegraphics[height=1cm]{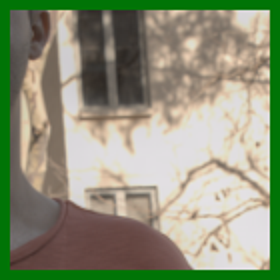} \vspace*{-6mm}
		\caption*{\tiny GT} \vspace*{-1.5mm}
		\caption*{\footnotesize }
	\end{subfigure}	
	\begin{subfigure}{.0554\textwidth}
		\includegraphics[height=1cm]{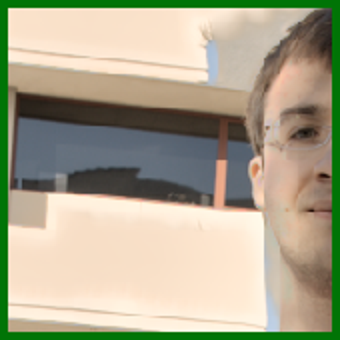} \vspace*{-6mm}
		\caption*{\tiny Sen} \vspace*{-1.5mm}
		\caption*{\footnotesize \cite{sen2012robust}}
	\end{subfigure}	
	\begin{subfigure}{.0554\textwidth}
		\includegraphics[height=1cm]{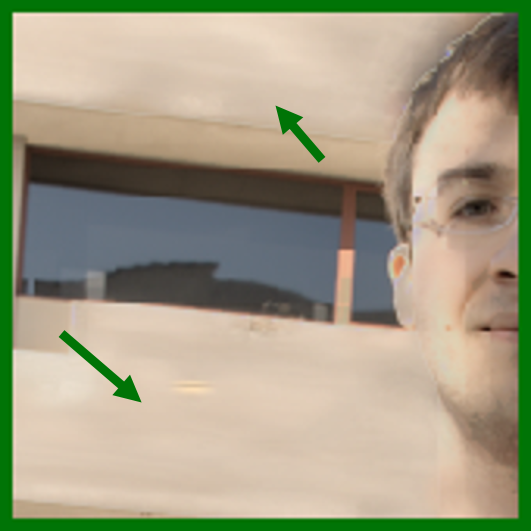} \vspace*{-6mm}
		\caption*{\tiny Hu} \vspace*{-1.5mm}
		\caption*{\footnotesize \cite{hu2013hdr}}
	\end{subfigure}	
	\begin{subfigure}{.0554\textwidth}
		\includegraphics[height=1cm]{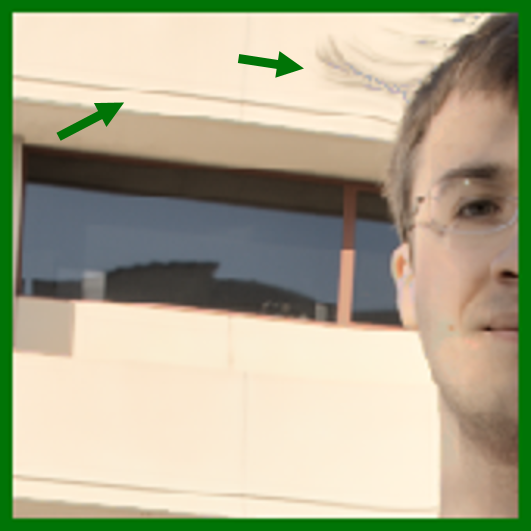} \vspace*{-6mm}
		\caption*{\tiny Kalantari} \vspace*{-1.5mm}
		\caption*{\footnotesize \cite{kalantari2017deep}}
	\end{subfigure}	
	\begin{subfigure}{.0554\textwidth}
		\includegraphics[height=1cm]{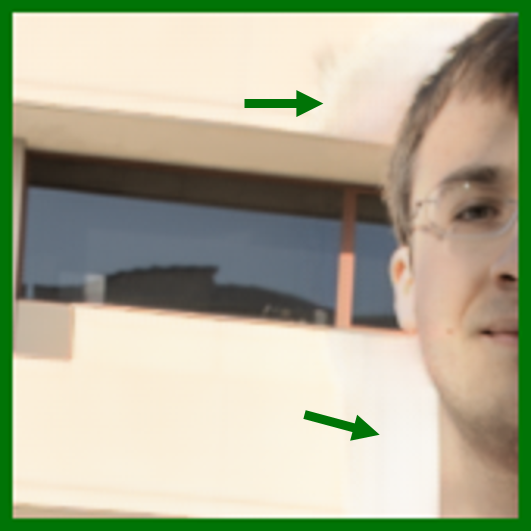} \vspace*{-6mm}
		\caption*{\tiny Wu} \vspace*{-1.5mm}
		\caption*{\footnotesize \cite{wu2018deep}}
	\end{subfigure}	
	\begin{subfigure}{.0554\textwidth}
		\includegraphics[height=1cm]{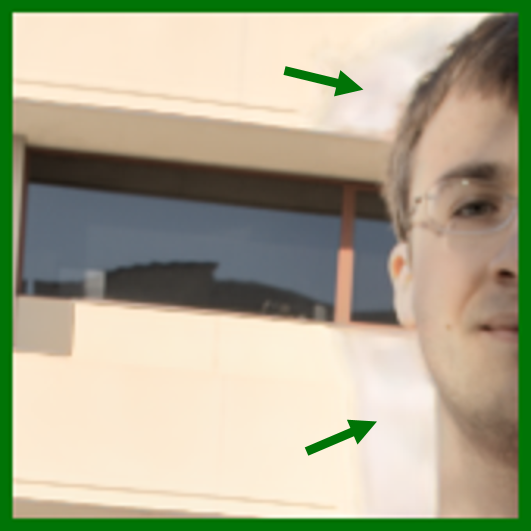} \vspace*{-6mm}
		\caption*{\tiny Yan} \vspace*{-1.5mm}
		\caption*{\footnotesize \cite{yan2019attention}}
	\end{subfigure}	
	\begin{subfigure}{.0554\textwidth}
		\includegraphics[height=1cm]{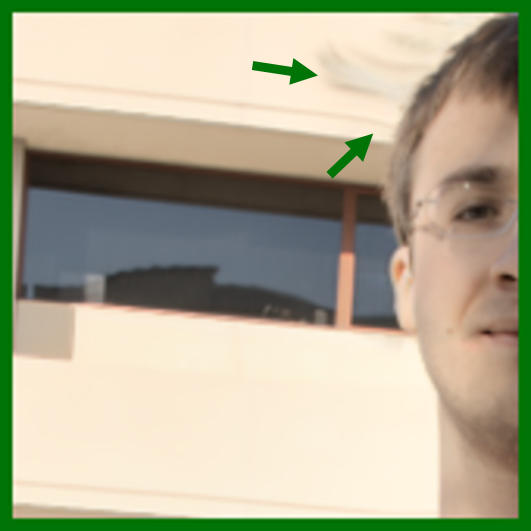} \vspace*{-6mm}
		\caption*{\tiny Prabhakar} \vspace*{-1.5mm}
		\caption*{\footnotesize \cite{Prabhakar_2021_CVPR}}
	\end{subfigure}	
	\begin{subfigure}{.0554\textwidth}
		\includegraphics[height=1cm]{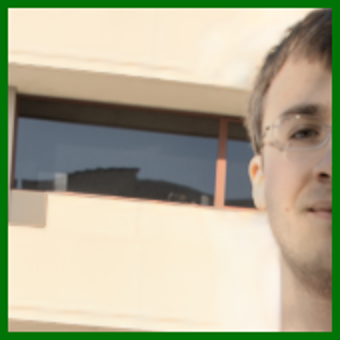} \vspace*{-6mm}
		\caption*{\tiny Ours} \vspace*{-1.5mm}
		\caption*{\footnotesize }
	\end{subfigure}	
	\begin{subfigure}{.0554\textwidth}
		\includegraphics[height=1cm]{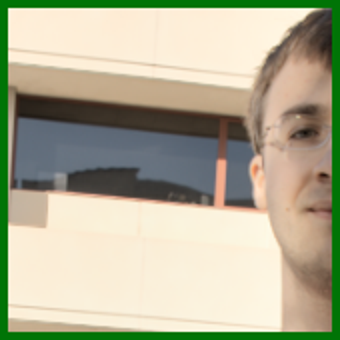} \vspace*{-6mm}
		\caption*{\tiny GT} \vspace*{-1.5mm}
		\caption*{\footnotesize }
	\end{subfigure}	

	\begin{subfigure}{.49\textwidth}
		\vspace*{1mm}
		\caption*{(a)} 
	\end{subfigure}	
	\begin{subfigure}{.49\textwidth}
		\vspace*{1mm}
		\caption*{(b)} 
	\end{subfigure}	
	\vspace*{-4mm}
	\caption{Qualitative comparisons of our method with state-of-the-art methods.}
	\label{fig:results}
\end{figure*}
	
\noindent\textbf{Qualitative evaluations} We compare our visual results with state-of-the-art methods. Fig.~\ref{fig:results} shows a challenging case where all the input images are over-exposed and large-scale motion exists. While the other methods fail to reconstruct detailed texture, our method successfully hallucinates details in the severely saturated regions. Our method also generates realistic details even when the given information is insufficient due to occlusions, while the other methods introduce ghosting artifacts or color distortions within the saturated areas.

\begin{figure}[t]
	\begin{subfigure}{.076\textwidth}
		\includegraphics[height=4.1cm]{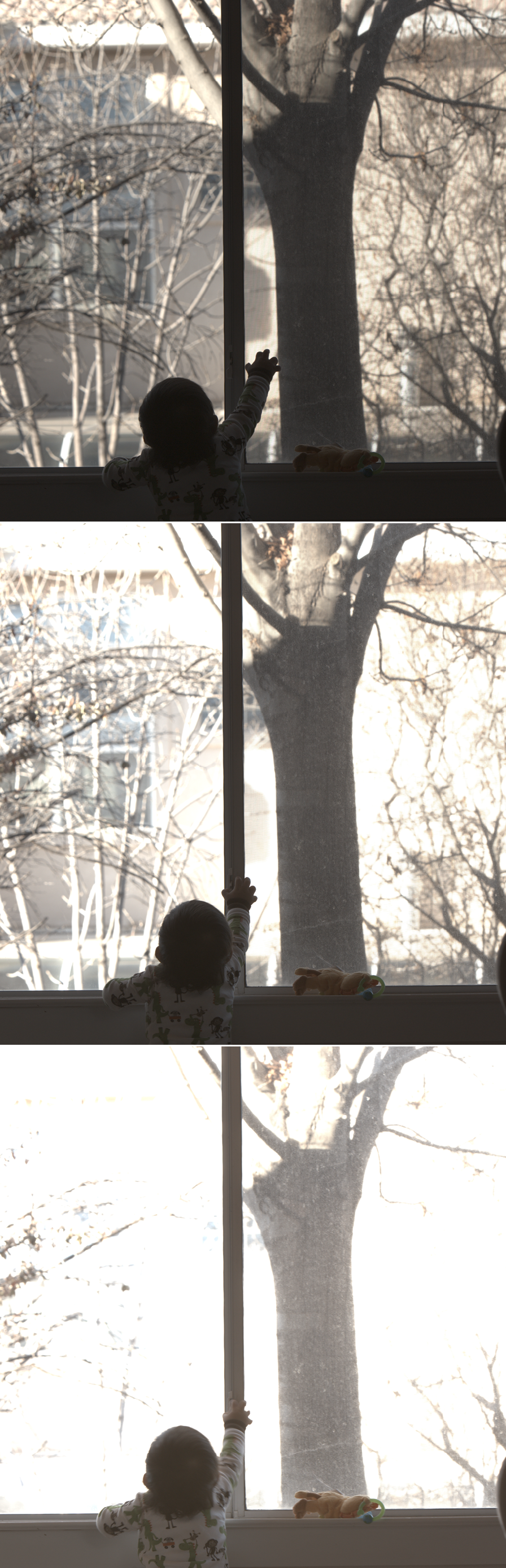} \vspace*{-5mm} \caption*{LDRs} 
	\end{subfigure}
	\begin{subfigure}{.228\textwidth}
		\includegraphics[height=4.1cm]{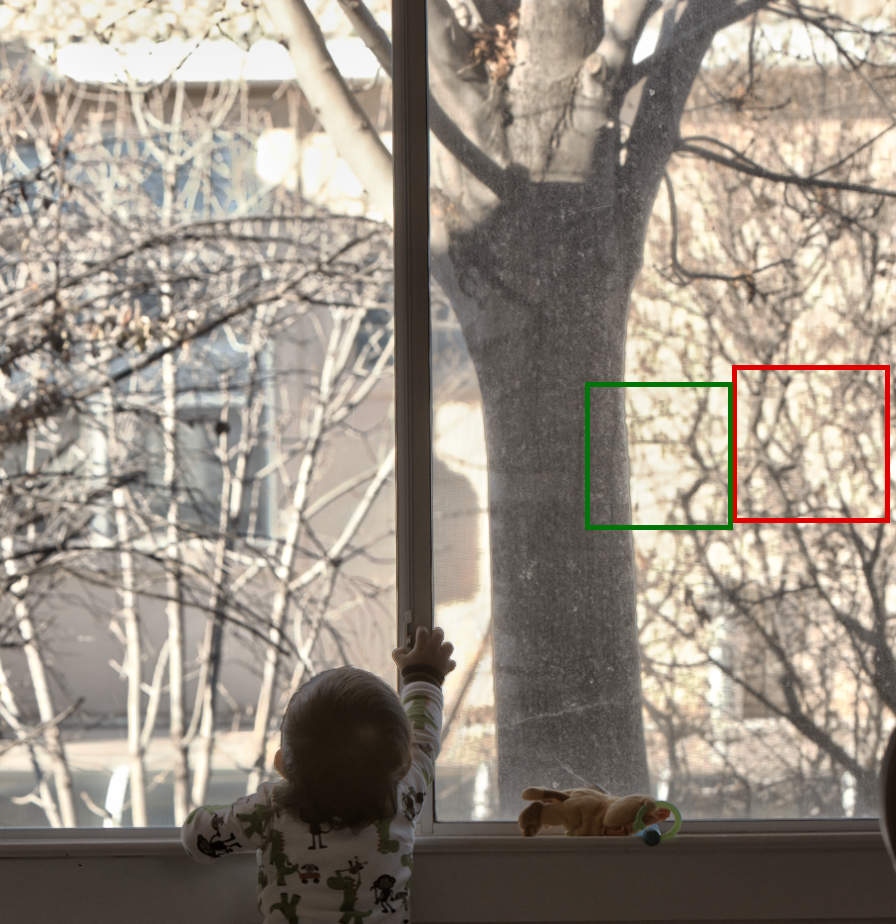} \vspace*{-5mm} \caption*{Our result}
	\end{subfigure}
	\begin{subfigure}{.076\textwidth}
		\includegraphics[height=4.1cm]{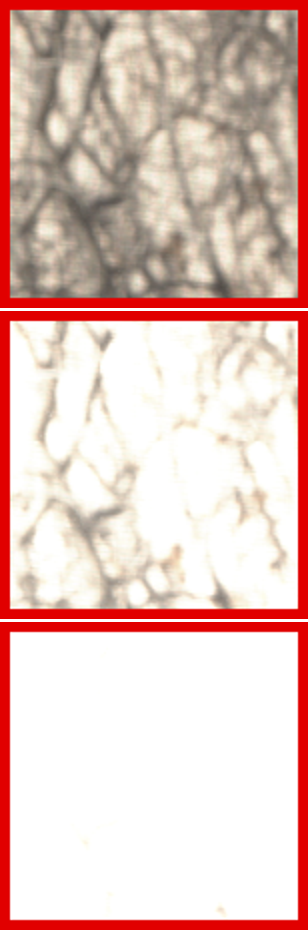} \vspace*{-5mm} \caption*{\quad\quad LDR}
	\end{subfigure}
	\begin{subfigure}{.056\textwidth}
		\includegraphics[height=4.1cm]{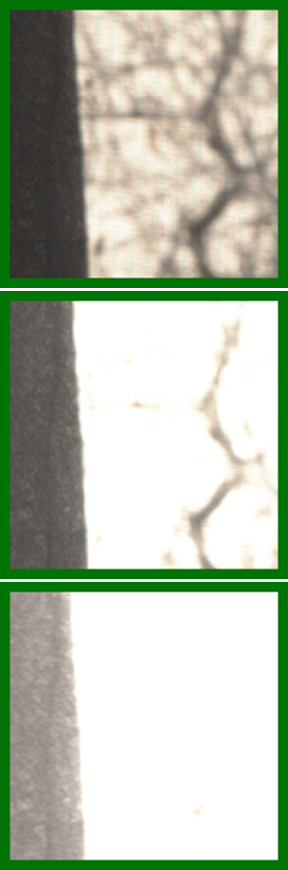} \vspace*{-5mm} \caption*{patches}
	\end{subfigure}	
	
	\begin{subfigure}{.0762\textwidth}
		\includegraphics[height=1.31cm]{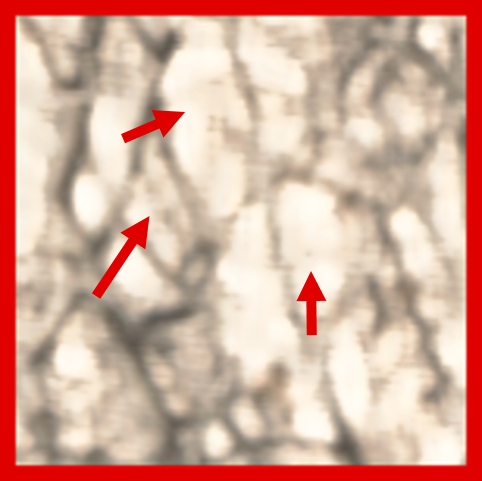} \vspace*{-5.5mm}
		\caption*{\footnotesize Yan} \vspace*{-1.5mm}
		\caption*{\footnotesize \cite{yan2019attention}}
	\end{subfigure}	
	\begin{subfigure}{.0762\textwidth}
		\includegraphics[height=1.31cm]{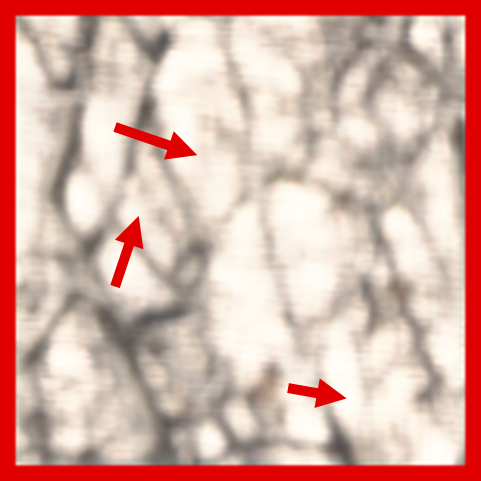} \vspace*{-5.5mm}
		\caption*{\footnotesize Prabhakar} \vspace*{-1.5mm}
		\caption*{\footnotesize \cite{Prabhakar_2021_CVPR}}
	\end{subfigure}	
	\begin{subfigure}{.0762\textwidth}
		\includegraphics[height=1.31cm]{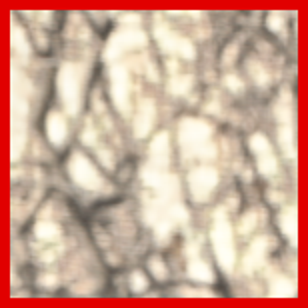} \vspace*{-5.5mm}
		\caption*{\footnotesize Ours} \vspace*{-1.5mm}
		\caption*{\footnotesize }
	\end{subfigure}		
	\begin{subfigure}{.0762\textwidth}
		\includegraphics[height=1.31cm]{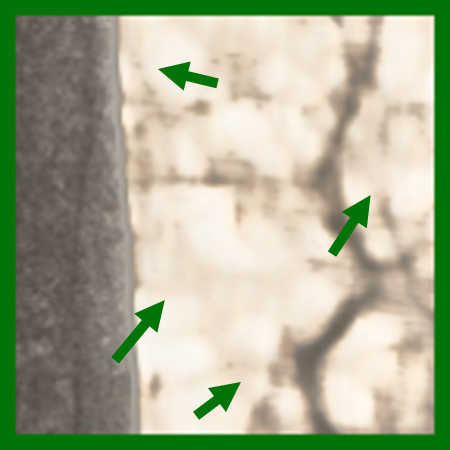} \vspace*{-5.5mm}
		\caption*{\footnotesize Yan} \vspace*{-1.5mm}
		\caption*{\footnotesize \cite{yan2019attention}}
	\end{subfigure}	
	\begin{subfigure}{.0762\textwidth}
		\includegraphics[height=1.31cm]{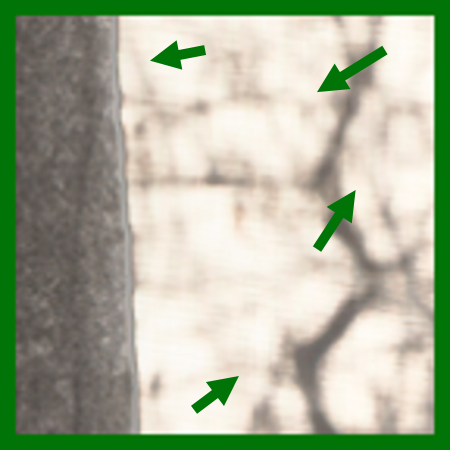} \vspace*{-5.5mm}
		\caption*{\footnotesize Prabhakar} \vspace*{-1.5mm}
		\caption*{\footnotesize \cite{Prabhakar_2021_CVPR}}
	\end{subfigure}	
	\begin{subfigure}{.0762\textwidth}
		\includegraphics[height=1.31cm]{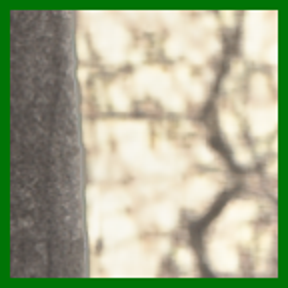} \vspace*{-5.5mm}
		\caption*{\footnotesize Ours} \vspace*{-1.5mm}
		\caption*{\footnotesize }
	\end{subfigure}

	\caption{Qualitative comparisons on the image from Sen \etal's \cite{sen2012robust} dataset.}
	\label{fig:result_sen}
\end{figure}

We also demonstrate our generalization ability by evaluating on Sen \etal’s dataset \cite{sen2012robust}. Fig.~\ref{fig:result_sen} shows the results on the case where the input images contain large saturated areas and especially the reference image provides little information. It can be shown that the proposed method successfully hallucinates details and texture in the saturated areas. More visual results are presented in the supplementary material.

\begin{table}[t]
	\caption{Comparisons of the proposed brightness adjustment strategy with the motion compensation scheme.}
	\label{tab:reformulation}
	\resizebox{\columnwidth}{!}{%
		\begin{tabular}{lcccc}
			\hline\\[-1em]
			Methods               & PSNR\textsubscript{$T$} & SSIM\textsubscript{$T$} & PSNR\textsubscript{$L$} & SSIM\textsubscript{$L$} \\ \hline\\[-1em]
			Motion Compensation   & 44.23         & 0.9913         & 42.21        & 0.9878       \\
			Brightness Adjustment & \textbf{44.48}  & \textbf{0.9917}  & \textbf{42.45}  & \textbf{0.9880}\\ \hline
		\end{tabular}%
	}
\end{table}

	\subsection{Ablation Study} \label{ablation}
	In this section, we evaluate the contributions of the proposed components.
	
\begin{figure}[t]
	\centering
	\begin{subfigure}{.42\textwidth}
		\centering
		\hspace{0.1mm}
		\includegraphics[width=.955\textwidth]{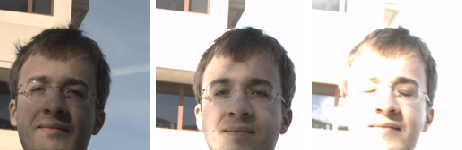}
		\vspace*{-0.5mm}
		\caption{Input LDR images}
		\vspace*{1mm}
	\end{subfigure}
	\begin{subfigure}{.13\textwidth}
		\includegraphics[width=\textwidth]{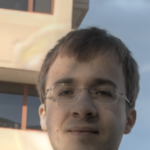}
		\vspace*{-4.5mm}
		\caption{M.C.}
	\end{subfigure}
	\begin{subfigure}{.13\textwidth}
		\includegraphics[width=\textwidth]{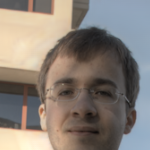}
		\vspace*{-4.5mm}
		\caption{B.A.}
	\end{subfigure}
	\begin{subfigure}{.13\textwidth}
		\includegraphics[width=\textwidth]{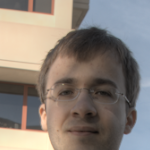}
		\vspace*{-4.5mm}
		\caption{GT}
	\end{subfigure}
	\caption{
		Effectiveness of the proposed brightness adjustment network. M.C. and B.A. denote motion compensation and brightness adjustment, respectively.}
	\label{fig:reformulation}
\end{figure}

	\noindent\textbf{Brightness adjustment} 
	The proposed approach generates well-aligned multi-exposure features by reformulating a motion alignment problem into a simple brightness adjustment problem. We demonstrate the effectiveness of this reformulation in Table~\ref{tab:reformulation}. We compare the proposed brightness adjustment method with the motion compensation method which uses the same architecture as the proposed one but aligns the supporting image to have the same structure as the reference as most previous works do. The motion compensation method needs to align the images with large motions and thus are prone to artifacts. Fig.~\ref{fig:reformulation} shows that the motion compensation scheme fails to handle the ghosting artifacts when the foreground motion exists in the over-exposed areas. It can be seen that the proposed reformulation strategy greatly eases the task and results in favorable performance. 

\begin{table}[t]
	\caption{Analysis on the MAHN architecture. Hall. and Refine. denote the hallucination branch and the refinement branch, respectively.}
	\label{tab:mahn}
	\resizebox{\columnwidth}{!}{%
		\begin{tabular}{ccccccc}
			\hline\\[-1em]
			\multicolumn{3}{c}{Methods}        & \multicolumn{1}{c}{\multirow{3}{*}{PSNR\textsubscript{$T$}}} & \multicolumn{1}{c}{\multirow{3}{*}{SSIM\textsubscript{$T$}}} & \multicolumn{1}{c}{\multirow{3}{*}{PSNR\textsubscript{$L$}}} & \multicolumn{1}{c}{\multirow{3}{*}{SSIM\textsubscript{$L$}}} \\ \cline{1-3}\\[-1em]
			Coarse  & \multicolumn{2}{c}{Fine} & \multicolumn{1}{c}{}                          & \multicolumn{1}{c}{}                          & \multicolumn{1}{c}{}                          & \multicolumn{1}{c}{}                           \\ \cline{2-3}\\[-1em]
					& Hall.      & Refine.     & \multicolumn{1}{c}{}                          & \multicolumn{1}{c}{}                          & \multicolumn{1}{c}{}                          & \multicolumn{1}{c}{}                           \\ \hline
			\checkmark  &              &         		& 41.58          & 0.9871          & 38.30    	& 0.9786          \\
			\checkmark  & \checkmark        	   &      & 44.17          & 0.9911   	   & 41.96           & 0.9869          \\
			\checkmark  & \checkmark   & \checkmark     & \textbf{44.48} & \textbf{0.9917} & \textbf{42.45}  & \textbf{0.9880}          \\ \hline
	\end{tabular}}
\end{table}
	\noindent\textbf{Coarse-to-fine MAHN} 
	Our MAHN first reconstructs a coarse HDR image and then hallucinates content in saturated regions in the fine network. We validate the effectiveness of our coarse-to-fine reconstruction strategy and contribution of each sub-network in Table~\ref{tab:mahn}. The coarse-to-fine architecture shows even better performances than the one-stage coarse network, even only with the hallucination branch. The proposed architecture including both the refinement branch and the hallucination branch produces the best results. It can be observed that the proposed coarse-to-fine architecture is effective and the hallucination branch greatly contributes to the HDR reconstruction process by performing explicit restoration for the saturated parts.
		
\begin{table}[t]
	\caption{Analysis on the soft adaptive contextual attention. $a$ adjusts the softness of the saturation mask $M$. The proposed method adopts $ a=3 $.}
	\label{tab:ca}
	\resizebox{\columnwidth}{!}{%
		\begin{tabular}{cccccc}
			\hline\\[-1em]
			\multicolumn{1}{l}{Methods}        & $a$ & PSNR\textsubscript{$T$} & SSIM\textsubscript{$T$} & PSNR\textsubscript{$L$} & SSIM\textsubscript{$L$} \\ \hline\\[-1em]
			\multicolumn{1}{l}{hard attention} &      & 43.47  & 0.9905  & 41.70   & 0.9863  \\ \hline\\[-1em]
			\multirow{3}{*}{soft attention}    & 0.5 & 43.61   & 0.9909  & 41.68  & 0.9868  \\
			& 1   & \textbf{44.52}  & 0.9915  & 42.34  & 0.9870  \\
			& 3   & 44.48  & \textbf{0.9917}  & \textbf{42.45}  & \textbf{0.9880} \\
			& 5   & 43.45  & 0.9906  & 41.32  & 0.9861  \\										   
			\hline
	\end{tabular}}
\end{table}

\newcommand{\wtext}{0.018}
\newcommand{\wfigures}{0.9}
\begin{figure}[t]
	\begin{minipage}{\wtext\textwidth}	
		\rotatebox[origin=c]{90}{\small \quad\qquad $H_{fine}$ \qquad\qquad $M$ \qquad\qquad $H_{coarse}$ \enspace} 
	\end{minipage}
\begin{minipage}{\wfigures\textwidth}	
	\begin{subfigure}{0.122\textwidth}
		\includegraphics[width=\textwidth]{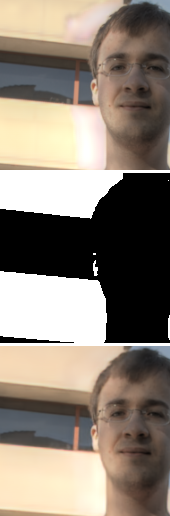}
		\caption*{Hard}
	\end{subfigure}
	\begin{subfigure}{0.122\textwidth}
		\includegraphics[width=\textwidth]{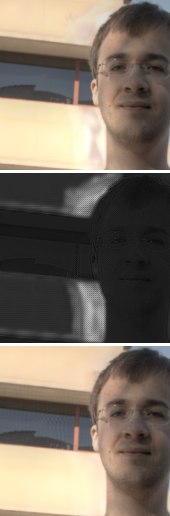}
		\caption*{$ a=1 $}
	\end{subfigure}
	\begin{subfigure}{0.122\textwidth}
		\includegraphics[width=\textwidth]{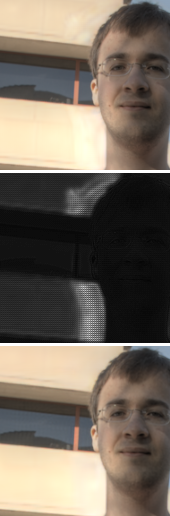}
		\caption*{$ a=3 $}
	\end{subfigure}
	\begin{subfigure}{0.122\textwidth}
		\includegraphics[width=\textwidth]{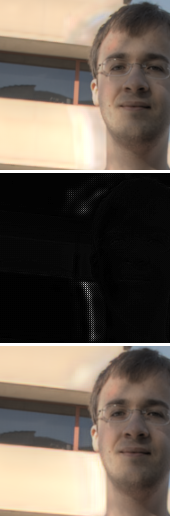}
		\caption*{$ a=5 $}
	\end{subfigure}
\end{minipage}
	\caption{Visual comparisons for different choices of the softness parameter $ a $ of the saturation mask $ M $. The proposed method adopts $ a=3 $.}
	\label{fig:softness}
\end{figure} 
		
	\noindent\textbf{Adaptive contextual attention} 
	We compensate for saturation using the soft adaptive contextual attention in the MAHN. To generate the saturation mask $ M $ representing the saturation level with values in range $ [0,1] $, we use a sigmoid function: $ sigmoid(x)={1}/(1+e^{-ax}) $, where $ a $ is a parameter that controls the steepness. We can adjust the softness of the saturation mask $M$ by changing the parameter $a$. As $ a $ increases, the resulting mask becomes closer to a hard (binary) mask as shown in Fig.~\ref{fig:softness}. When $a$ is small, the mask becomes too soft to clearly identify the saturated regions which need to be restored.
	On the other hand, when $a$ is too big, the value should be almost $0$ or $1$, thus it is difficult to reconstruct smooth and natural images. Table~\ref{tab:ca} shows the results with different softness parameter $a$. The best performances are achieved with the modest softness $a=3$.
	We also compare the proposed soft adaptive attention with hard attention which is similar to the original contextual attention. The hard attention method generates the saturation mask by thresholding the coarse HDR image $H_{coarse}$ with a threshold of $\tau$: $M(x,y) = \mathbbm{1}[ H_{coarse}(x,y) \ge \tau ]$, where $(x,y)$ denotes a pixel location and $\mathbbm{1}$ is an indicator function. $\tau$ is set as $0.9$ empirically. 
	We observe that the soft adaptive contextual attention enables sample-specific mask generation and hallucination, which leads to detailed HDR image generation. 	
	\section{Conclusion}
We have proposed an end-to-end HDR imaging CNN, which takes multi-exposure inputs with dynamic motions and generates ghost-free HDR images with some hallucinations in washed-out regions.
For this, we have introduced the BAN which adjusts the brightness of the reference feature using adaptive convolutions so that the well-aligned multi-exposure features are generated. Then, the bracketed features are integrated into a clean HDR image with supervision on the saturated regions. We have also proposed the MAHN, which reconstructs details in the saturated areas by aggregating valid content from the unsaturated regions. Experiments show that the proposed system delivers high-quality HDR results even in the presence of severe saturation and large displacement. Our code is available at 
\url{https://github.com/haesoochung/hdri-saturation-compensation}.

\section*{Acknowledgments}
This work was supported in part by the National Research Foundation of Korea(NRF) grant funded by the Korea government(MSIT) (2021R1A2C2007220), and in part by Samsung Electronics Co., Ltd.

	\newpage
	
	{\small
		\bibliographystyle{ieee_fullname}
		\bibliography{bib}

\begin{thebibliography}{10}\itemsep=-1pt

\bibitem{akyuz2007hdr}
Ahmet~Oǧuz Aky{\"u}z, Roland Fleming, Bernhard~E Riecke, Erik Reinhard, and
  Heinrich~H B{\"u}lthoff.
\newblock Do hdr displays support ldr content? a psychophysical evaluation.
\newblock {\em ACM Transactions on Graphics (TOG)}, 26(3):38--es, 2007.

\bibitem{an2012reduction}
Jaehyun An, Seong~Jong Ha, and Nam~Ik Cho.
\newblock Reduction of ghost effect in exposure fusion by detecting the ghost
  pixels in saturated and non-saturated regions.
\newblock In {\em 2012 IEEE International Conference on Acoustics, Speech and
  Signal Processing (ICASSP)}, pages 1101--1104. IEEE, 2012.

\bibitem{an2014probabilistic}
Jaehyun An, Seong~Jong Ha, and Nam~Ik Cho.
\newblock Probabilistic motion pixel detection for the reduction of ghost
  artifacts in high dynamic range images from multiple exposures.
\newblock {\em EURASIP Journal on Image and Video Processing}, 2014(1):1--15,
  2014.

\bibitem{an2011multi}
Jaehyun An, Sang~Heon Lee, Jung~Gap Kuk, and Nam~Ik Cho.
\newblock A multi-exposure image fusion algorithm without ghost effect.
\newblock In {\em 2011 IEEE International Conference on Acoustics, Speech and
  Signal Processing (ICASSP)}, pages 1565--1568. IEEE, 2011.

\bibitem{banterle2006inverse}
Francesco Banterle, Patrick Ledda, Kurt Debattista, and Alan Chalmers.
\newblock Inverse tone mapping.
\newblock In {\em Proceedings of the 4th international conference on Computer
  graphics and interactive techniques in Australasia and Southeast Asia}, pages
  349--356, 2006.

\bibitem{banterle2007framework}
Francesco Banterle, Patrick Ledda, Kurt Debattista, Alan Chalmers, and Marina
  Bloj.
\newblock A framework for inverse tone mapping.
\newblock {\em The Visual Computer}, 23(7):467--478, 2007.

\bibitem{bertasius2018object}
Gedas Bertasius, Lorenzo Torresani, and Jianbo Shi.
\newblock Object detection in video with spatiotemporal sampling networks.
\newblock In {\em Proceedings of the European Conference on Computer Vision
  (ECCV)}, pages 331--346, 2018.

\bibitem{dai2017deformable}
Jifeng Dai, Haozhi Qi, Yuwen Xiong, Yi Li, Guodong Zhang, Han Hu, and Yichen
  Wei.
\newblock Deformable convolutional networks.
\newblock In {\em Proceedings of the IEEE international conference on computer
  vision}, pages 764--773, 2017.

\bibitem{debevec1997recovering}
Paul~E Debevec and Jitendra Malik.
\newblock Recovering high dynamic range radiance maps from photographs.
\newblock In {\em Proceedings of the 24th annual conference on Computer
  graphics and interactive techniques}, pages 369--378. ACM
  Press/Addison-Wesley Publishing Co., 1997.

\bibitem{didyk2008enhancement}
Piotr Didyk, Rafal Mantiuk, Matthias Hein, and Hans-Peter Seidel.
\newblock Enhancement of bright video features for hdr displays.
\newblock In {\em Computer Graphics Forum}, volume~27, pages 1265--1274. Wiley
  Online Library, 2008.

\bibitem{eilertsen2017hdr}
Gabriel Eilertsen, Joel Kronander, Gyorgy Denes, Rafa{\l}~K Mantiuk, and Jonas
  Unger.
\newblock Hdr image reconstruction from a single exposure using deep cnns.
\newblock {\em ACM Transactions on Graphics (TOG)}, 36(6):178, 2017.

\bibitem{endoSA2017}
Yuki Endo, Yoshihiro Kanamori, and Jun Mitani.
\newblock Deep reverse tone mapping.
\newblock {\em ACM Transactions on Graphics (Proc. of SIGGRAPH ASIA 2017)},
  36(6), Nov. 2017.

\bibitem{grosch2006fast}
Thorsten Grosch et~al.
\newblock Fast and robust high dynamic range image generation with camera and
  object movement.
\newblock {\em Vision, Modeling and Visualization, RWTH Aachen}, 277284, 2006.

\bibitem{heo2010ghost}
Yong~Seok Heo, Kyoung~Mu Lee, Sang~Uk Lee, Youngsu Moon, and Joonhyuk Cha.
\newblock Ghost-free high dynamic range imaging.
\newblock In {\em Asian Conference on Computer Vision}, pages 486--500.
  Springer, 2010.

\bibitem{hu2013hdr}
Jun Hu, Orazio Gallo, Kari Pulli, and Xiaobai Sun.
\newblock Hdr deghosting: How to deal with saturation?
\newblock In {\em Proceedings of the IEEE Conference on Computer Vision and
  Pattern Recognition}, pages 1163--1170, 2013.

\bibitem{huo2014physiological}
Yongqing Huo, Fan Yang, Le Dong, and Vincent Brost.
\newblock Physiological inverse tone mapping based on retina response.
\newblock {\em The Visual Computer}, 30(5):507--517, 2014.

\bibitem{jia2016dynamic}
Xu Jia, Bert De~Brabandere, Tinne Tuytelaars, and Luc Van~Gool.
\newblock Dynamic filter networks.
\newblock In {\em NIPS}, 2016.

\bibitem{kalantari2017deep}
Nima~Khademi Kalantari and Ravi Ramamoorthi.
\newblock Deep high dynamic range imaging of dynamic scenes.
\newblock {\em ACM Trans. Graph.}, 36(4):144--1, 2017.

\bibitem{kang2003high}
Sing~Bing Kang, Matthew Uyttendaele, Simon Winder, and Richard Szeliski.
\newblock High dynamic range video.
\newblock In {\em ACM Transactions on Graphics (TOG)}, volume~22, pages
  319--325. ACM, 2003.

\bibitem{khan2006ghost}
Erum~Arif Khan, Ahmet~Oguz Akyuz, and Erik Reinhard.
\newblock Ghost removal in high dynamic range images.
\newblock In {\em 2006 International Conference on Image Processing}, pages
  2005--2008. IEEE, 2006.

\bibitem{kim2018spatio}
Tae~Hyun Kim, Mehdi~SM Sajjadi, Michael Hirsch, and Bernhard Scholkopf.
\newblock Spatio-temporal transformer network for video restoration.
\newblock In {\em Proceedings of the European Conference on Computer Vision
  (ECCV)}, pages 106--122, 2018.

\bibitem{kovaleski2009high}
Rafael~Pacheco Kovaleski and Manuel~M Oliveira.
\newblock High-quality brightness enhancement functions for real-time reverse
  tone mapping.
\newblock {\em The Visual Computer}, 25(5):539--547, 2009.

\bibitem{kovaleski2014high}
Rafael~P Kovaleski and Manuel~M Oliveira.
\newblock High-quality reverse tone mapping for a wide range of exposures.
\newblock In {\em 2014 27th SIBGRAPI Conference on Graphics, Patterns and
  Images}, pages 49--56. IEEE, 2014.

\bibitem{landis2002production}
Hayden Landis.
\newblock Production-ready global illumination.
\newblock In {\em Siggraph 2002}, volume~5, pages 93--95, 2002.

\bibitem{lee2014ghost}
Chul Lee, Yuelong Li, and Vishal Monga.
\newblock Ghost-free high dynamic range imaging via rank minimization.
\newblock {\em IEEE Signal Processing Letters}, 21(9):1045--1049, 2014.

\bibitem{lee2018deeprecursive}
Siyeong Lee, Gwon~Hwan An, and Suk-Ju Kang.
\newblock Deep recursive hdri: Inverse tone mapping using generative
  adversarial networks.
\newblock In {\em Proceedings of the European Conference on Computer Vision
  (ECCV)}, pages 596--611, 2018.

\bibitem{liu2020single}
Yu-Lun Liu, Wei-Sheng Lai, Yu-Sheng Chen, Yi-Lung Kao, Ming-Hsuan Yang, Yung-Yu
  Chuang, and Jia-Bin Huang.
\newblock Single-image hdr reconstruction by learning to reverse the camera
  pipeline.
\newblock In {\em Proceedings of the IEEE/CVF Conference on Computer Vision and
  Pattern Recognition}, pages 1651--1660, 2020.

\bibitem{mann1994beingundigital}
S Mann and R Picard.
\newblock Beingundigital’with digital cameras.
\newblock {\em MIT Media Lab Perceptual}, 1:2, 1994.

\bibitem{mantiuk2011hdr}
Rafat Mantiuk, Kil~Joong Kim, Allan~G Rempel, and Wolfgang Heidrich.
\newblock Hdr-vdp-2: a calibrated visual metric for visibility and quality
  predictions in all luminance conditions.
\newblock In {\em ACM Transactions on graphics (TOG)}, volume~30, page~40. ACM,
  2011.

\bibitem{marnerides2018expandnet}
Demetris Marnerides, Thomas Bashford-Rogers, Jonathan Hatchett, and Kurt
  Debattista.
\newblock Expandnet: A deep convolutional neural network for high dynamic range
  expansion from low dynamic range content.
\newblock In {\em Computer Graphics Forum}, volume~37, pages 37--49. Wiley
  Online Library, 2018.

\bibitem{meylan2006reproduction}
Laurence Meylan, Scott Daly, and Sabine S{\"u}sstrunk.
\newblock The reproduction of specular highlights on high dynamic range
  displays.
\newblock In {\em Color and Imaging Conference}, volume 2006, pages 333--338.
  Society for Imaging Science and Technology, 2006.

\bibitem{mildenhall2018burst}
Ben Mildenhall, Jonathan~T Barron, Jiawen Chen, Dillon Sharlet, Ren Ng, and
  Robert Carroll.
\newblock Burst denoising with kernel prediction networks.
\newblock In {\em Proceedings of the IEEE Conference on Computer Vision and
  Pattern Recognition}, pages 2502--2510, 2018.

\bibitem{niklaus2017video}
Simon Niklaus, Long Mai, and Feng Liu.
\newblock Video frame interpolation via adaptive convolution.
\newblock In {\em Proceedings of the IEEE Conference on Computer Vision and
  Pattern Recognition}, pages 670--679, 2017.

\bibitem{niu2021hdr}
Yuzhen Niu, Jianbin Wu, Wenxi Liu, Wenzhong Guo, and Rynson~WH Lau.
\newblock Hdr-gan: Hdr image reconstruction from multi-exposed ldr images with
  large motions.
\newblock {\em IEEE Transactions on Image Processing}, 30:3885--3896, 2021.

\bibitem{oh2015robust}
Tae-Hyun Oh, Joon-Young Lee, Yu-Wing Tai, and In~So Kweon.
\newblock Robust high dynamic range imaging by rank minimization.
\newblock {\em IEEE transactions on pattern analysis and machine intelligence},
  37(6):1219--1232, 2015.

\bibitem{prabhakartowards}
K~Ram Prabhakar, Susmit Agrawal, Durgesh~Kumar Singh, Balraj Ashwath, and
  R~Venkatesh Babu.
\newblock Towards practical and efficient high-resolution hdr deghosting with
  cnn.

\bibitem{prabhakar2016ghosting}
K~Ram Prabhakar and R~Venkatesh Babu.
\newblock Ghosting-free multi-exposure image fusion in gradient domain.
\newblock In {\em 2016 IEEE International Conference on Acoustics, Speech and
  Signal Processing (ICASSP)}, pages 1766--1770. IEEE, 2016.

\bibitem{Prabhakar_2021_CVPR}
K.~Ram Prabhakar, Gowtham Senthil, Susmit Agrawal, R.~Venkatesh Babu, and Rama
  Krishna Sai~S Gorthi.
\newblock Labeled from unlabeled: Exploiting unlabeled data for few-shot deep
  hdr deghosting.
\newblock In {\em Proceedings of the IEEE/CVF Conference on Computer Vision and
  Pattern Recognition (CVPR)}, pages 4875--4885, June 2021.

\bibitem{pu2020robust}
Zhiyuan Pu, Peiyao Guo, M~Salman Asif, and Zhan Ma.
\newblock Robust high dynamic range (hdr) imaging with complex motion and
  parallax.
\newblock In {\em Proceedings of the Asian Conference on Computer Vision},
  2020.

\bibitem{raman2011reconstruction}
Shanmuganathan Raman and Subhasis Chaudhuri.
\newblock Reconstruction of high contrast images for dynamic scenes.
\newblock {\em The Visual Computer}, 27(12):1099--1114, 2011.

\bibitem{rempel2007ldr2hdr}
Allan~G Rempel, Matthew Trentacoste, Helge Seetzen, H~David Young, Wolfgang
  Heidrich, Lorne Whitehead, and Greg Ward.
\newblock Ldr2hdr: on-the-fly reverse tone mapping of legacy video and
  photographs.
\newblock {\em ACM transactions on graphics (TOG)}, 26(3):39--es, 2007.

\bibitem{sen2012robust}
Pradeep Sen, Nima~Khademi Kalantari, Maziar Yaesoubi, Soheil Darabi, Dan~B
  Goldman, and Eli Shechtman.
\newblock Robust patch-based hdr reconstruction of dynamic scenes.
\newblock {\em ACM Trans. Graph.}, 31(6):203--1, 2012.

\bibitem{tian2020tdan}
Yapeng Tian, Yulun Zhang, Yun Fu, and Chenliang Xu.
\newblock Tdan: Temporally-deformable alignment network for video
  super-resolution.
\newblock In {\em Proceedings of the IEEE/CVF Conference on Computer Vision and
  Pattern Recognition}, pages 3360--3369, 2020.

\bibitem{wang2019edvr}
Xintao Wang, Kelvin~CK Chan, Ke Yu, Chao Dong, and Chen Change~Loy.
\newblock Edvr: Video restoration with enhanced deformable convolutional
  networks.
\newblock In {\em Proceedings of the IEEE/CVF Conference on Computer Vision and
  Pattern Recognition Workshops}, pages 0--0, 2019.

\bibitem{wu2018deep}
Shangzhe Wu, Jiarui Xu, Yu-Wing Tai, and Chi-Keung Tang.
\newblock Deep high dynamic range imaging with large foreground motions.
\newblock In {\em Proceedings of the European Conference on Computer Vision
  (ECCV)}, pages 117--132, 2018.

\bibitem{xiang2020zooming}
Xiaoyu Xiang, Yapeng Tian, Yulun Zhang, Yun Fu, Jan~P Allebach, and Chenliang
  Xu.
\newblock Zooming slow-mo: Fast and accurate one-stage space-time video
  super-resolution.
\newblock In {\em Proceedings of the IEEE/CVF conference on computer vision and
  pattern recognition}, pages 3370--3379, 2020.

\bibitem{yan2019attention}
Qingsen Yan, Dong Gong, Qinfeng Shi, Anton van~den Hengel, Chunhua Shen, Ian
  Reid, and Yanning Zhang.
\newblock Attention-guided network for ghost-free high dynamic range imaging.
\newblock {\em arXiv preprint arXiv:1904.10293}, 2019.

\bibitem{yan2019multi}
Qingsen Yan, Dong Gong, Pingping Zhang, Qinfeng Shi, Jinqiu Sun, Ian Reid, and
  Yanning Zhang.
\newblock Multi-scale dense networks for deep high dynamic range imaging.
\newblock In {\em 2019 IEEE Winter Conference on Applications of Computer
  Vision (WACV)}, pages 41--50. IEEE, 2019.

\bibitem{yan2020deep}
Qingsen Yan, Lei Zhang, Yu Liu, Yu Zhu, Jinqiu Sun, Qinfeng Shi, and Yanning
  Zhang.
\newblock Deep hdr imaging via a non-local network.
\newblock {\em IEEE Transactions on Image Processing}, 29:4308--4322, 2020.

\bibitem{Yang_2018_CVPR}
Xin Yang, Ke Xu, Yibing Song, Qiang Zhang, Xiaopeng Wei, and Rynson~W.H. Lau.
\newblock Image correction via deep reciprocating hdr transformation.
\newblock In {\em Proceedings of the IEEE Conference on Computer Vision and
  Pattern Recognition (CVPR)}, June 2018.

\bibitem{ying2020deformable}
Xinyi Ying, Longguang Wang, Yingqian Wang, Weidong Sheng, Wei An, and Yulan
  Guo.
\newblock Deformable 3d convolution for video super-resolution.
\newblock {\em IEEE Signal Processing Letters}, 27:1500--1504, 2020.

\bibitem{yu2018generative}
Jiahui Yu, Zhe Lin, Jimei Yang, Xiaohui Shen, Xin Lu, and Thomas~S Huang.
\newblock Generative image inpainting with contextual attention.
\newblock In {\em Proceedings of the IEEE conference on computer vision and
  pattern recognition}, pages 5505--5514, 2018.

\bibitem{zhang2017learning}
Jinsong Zhang and Jean-Fran{\c{c}}ois Lalonde.
\newblock Learning high dynamic range from outdoor panoramas.
\newblock In {\em Proceedings of the IEEE International Conference on Computer
  Vision}, pages 4519--4528, 2017.

\bibitem{zhang2012gradient}
Wei Zhang and Wai-Kuen Cham.
\newblock Gradient-directed multiexposure composition.
\newblock {\em IEEE Transactions on Image Processing}, 21(4):2318--2323, 2012.

\bibitem{zhao2018trajectory}
Yue Zhao, Yuanjun Xiong, and Dahua Lin.
\newblock Trajectory convolution for action recognition.
\newblock In {\em Proceedings of the 32nd International Conference on Neural
  Information Processing Systems}, pages 2208--2219, 2018.

\bibitem{zhu2019deformable}
Xizhou Zhu, Han Hu, Stephen Lin, and Jifeng Dai.
\newblock Deformable convnets v2: More deformable, better results.
\newblock In {\em Proceedings of the IEEE/CVF Conference on Computer Vision and
  Pattern Recognition}, pages 9308--9316, 2019.

\bibitem{zimmer2011freehand}
Henning Zimmer, Andr{\'e}s Bruhn, and Joachim Weickert.
\newblock Freehand hdr imaging of moving scenes with simultaneous resolution
  enhancement.
\newblock In {\em Computer Graphics Forum}, volume~30, pages 405--414. Wiley
  Online Library, 2011.

\end{thebibliography}
	}
	
\end{document}